\documentclass[10pt,journal,compsoc]{IEEEtran}

\usepackage{graphicx}
\usepackage{tikz}
\usepackage{comment}
\usepackage{amsmath,amssymb} % define this before the line numbering.
\usepackage{color}
\usepackage{multirow}
\usepackage{dsfont}
\usepackage{caption}
\usepackage{subcaption}

\definecolor{ultramarine}{RGB}{18, 10, 143}
\definecolor{darkred}{RGB}{187, 39, 26}
\definecolor{green}{RGB}{0,128,0}
\usepackage[colorlinks=true,linkcolor=red,citecolor=ultramarine]{hyperref}  % hyperlinks
\newcommand{\blue}[1]{\textcolor{blue}{#1}}
\newcommand{\red}[1]{\textcolor{red}{#1}}
\newcommand{\darkred}[1]{\textcolor{darkred}{#1}}

\newcommand{\ie}{\textit{i}.\textit{e}.}
\newcommand{\eg}{\textit{e}.\textit{g}.}

\def\model{Latent Anchor}

\ifCLASSOPTIONcompsoc
  \usepackage[nocompress]{cite}
\else
  \usepackage{cite} 
\fi

\ifCLASSINFOpdf
\else
\fi

\begin{document}

\title{Domain-Scalable Unpaired Image Translation via Latent Space Anchoring}

\author{Siyu Huang*, Jie An*, Donglai Wei, Zudi Lin,
        Jiebo Luo,~\IEEEmembership{Fellow,~IEEE}, 
        and Hanspeter Pfister,~\IEEEmembership{Fellow,~IEEE}% <-this % stops a space
\IEEEcompsocitemizethanks{
\IEEEcompsocthanksitem *: Equal contribution
\IEEEcompsocthanksitem Siyu Huang, Zudi Lin, and Hanspeter Pfister (corresponding author) are with the John A. Paulson School of Engineering and Applied Sciences, Harvard University, Boston, MA 02134. E-mail: \{huang,linzudi,pfister\}@seas.harvard.edu. Zudi Lin is presently affiliated with Amazon Alexa Science and the work was done before joining Amazon.
% \protect\\
\IEEEcompsocthanksitem Jie An and Jiebo Luo are with the Department of Computer Science, University or Rochester, Rochester, NY 14627. E-mail: \{jan6,jluo\}@cs.rochester.edu
\IEEEcompsocthanksitem Donglai Wei is with the Computer Science Department, Boston College, Boston, MA 02116. E-mail: donglai.wei@bc.edu
}% <-this % stops an unwanted space
\thanks{Manuscript 
received August 4, 2022;
% revised December 11, 2022 and April 26, 2023; 
accepted June 4, 2023. 
}
}

\markboth{IEEE Transactions on Pattern Analysis and Machine Intelligence} {Huang \MakeLowercase{\textit{et al.}}: Domain-Scalable Unpaired Image Translation via Latent Space Anchoring}

\IEEEtitleabstractindextext{%
\begin{abstract}
Unpaired image-to-image translation (UNIT) aims to map images between two visual domains without paired training data. However, given a UNIT model trained on certain domains, it is difficult for current methods to incorporate new domains because they often need to train the full model on both existing and new domains. To address this problem, we propose a new domain-scalable UNIT method, termed as \emph{latent space anchoring}, which can be efficiently extended to new visual domains and does not need to fine-tune encoders and decoders of existing domains. Our method anchors images of different domains to the same latent space of frozen GANs by learning lightweight encoder and regressor models to reconstruct single-domain images. In the inference phase, the learned encoders and decoders of different domains can be arbitrarily combined to translate images between any two domains without fine-tuning. Experiments on various datasets show that the proposed method achieves superior performance on both standard and domain-scalable UNIT tasks in comparison with the state-of-the-art methods. 
\end{abstract}

}

\maketitle

\IEEEdisplaynontitleabstractindextext

\IEEEpeerreviewmaketitle

\IEEEraisesectionheading{\section{Introduction}\label{sec:introduction}}

\begin{figure}[t]
    \centering
    \includegraphics[width=1\linewidth]{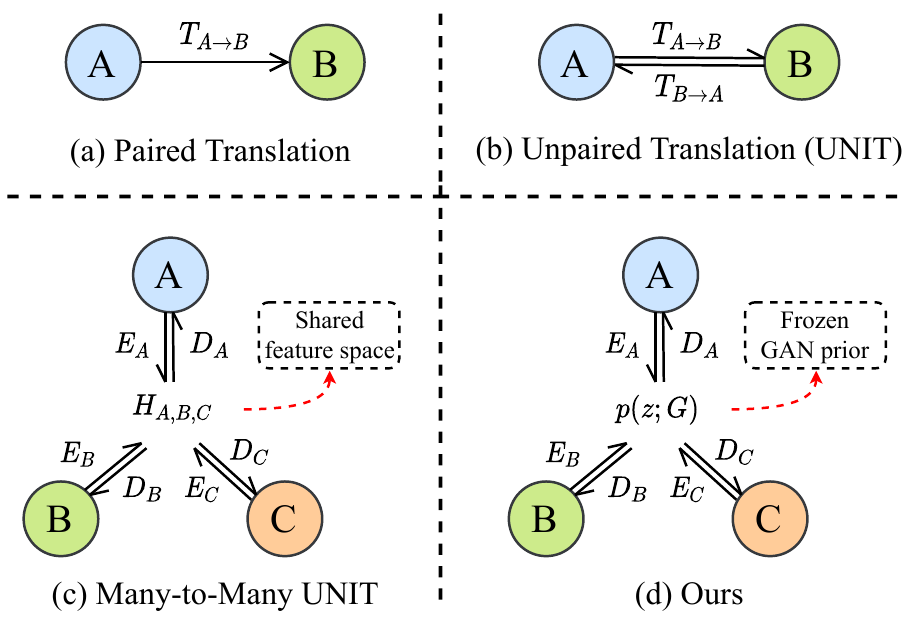}
    \caption{An illustration of the image-to-image translation paradigms. 
    \textbf{(a)} Dual-domain paired image translation, \eg, Pix2Pix~\cite{pix2pix} and SPADE \cite{spade}, learns a one-way mapping $T_{A \rightarrow B}$ from domain $A$ to domain $B$.
    \textbf{(b)} Dual-domain unpaired image translation, \eg, CycleGAN~\cite{cyclegan} and UNIT \cite{unit}, learns a bi-directional mapping $T_{A \rightarrow B}$ and $T_{B \rightarrow A}$ between two domains.
    \textbf{(c)} Many-to-many unpaired image translation, \eg, StarGAN~\cite{cyclegan} and DRIT++ \cite{drit++}, learns mappings between any two domains via a shared feature space $H$, encoders $E_{\{A,B,C\}}$, and decoders $D_{\{A,B,C\}}$.
    \textbf{(d)} Our latent space anchoring method learns mappings between any two domains via the latent space $p(z)$ of a frozen GAN $G$.
    }
    \label{fig:task}
\end{figure}

\IEEEPARstart{I}{mage-to-image} translation aims to map images from one visual domain to another (Fig.~\ref{fig:task}\red{a}),
where visual domains can be different image modalities (\eg, RGB $\leftrightarrow$ sketch), styles (\eg, winter $\leftrightarrow$ summer), or objects (\eg, horse $\leftrightarrow$ zebra).
Due to the scarcity of paired data, unsupervised image-to-image translation (UNIT) methods ~\cite{cyclegan,unit,cut} have been proposed to learn a bi-directional mapping between image domains (Fig.~\ref{fig:task}\red{b}). 
However, it is challenging to extend UNIT into more than two domains. Multi-domain UNIT methods are thus proposed to avoid the complexity of learning translators for all pairs of domains, which maintain a shared latent space for all available domains (Fig.~\ref{fig:task}\red{c}) while an encoder and a decoder of each domain are trained to map images to the shared feature space and invert features back to images, respectively. 

\begin{figure}[t]
    \centering
    \hspace{-1.4em}
    \includegraphics[width=1.04\linewidth]{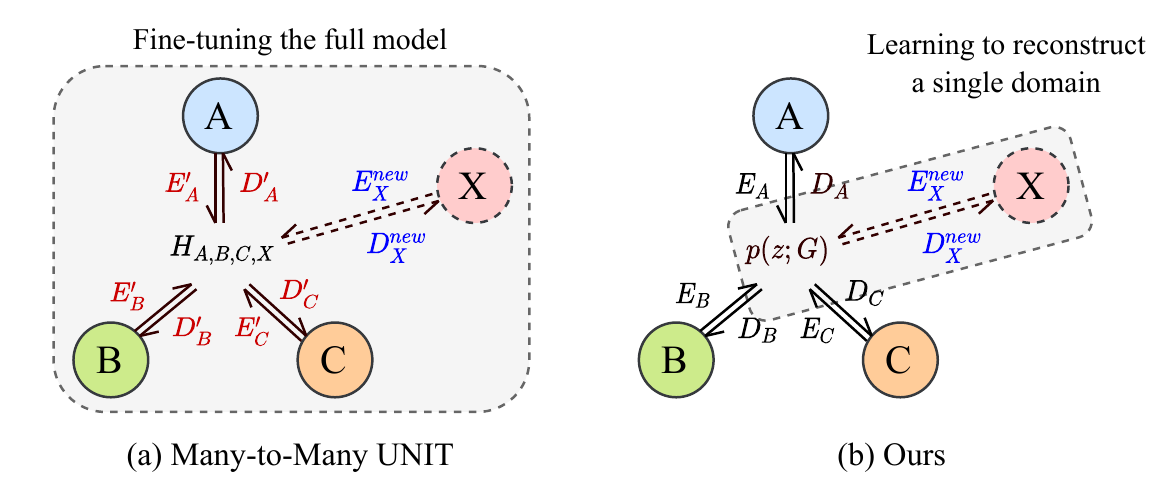}
    \caption{Adding a new domain $X$ to a learned UNIT model. \darkred{Red}: Models that need to be fine-tuned. \blue{Blue}: Models that need to be trained from scratch. \textbf{(a)} Existing methods usually need to fine-tune the full model on data of all available image domains. \textbf{(b)} Our method only needs to learn an encoder $E_{X}^{new}$ and a decoder $D_{X}^{new}$ on data of the new domain $X$, then images of the domain $X$ can be translated from/to previous domains via the latent space of the generator.
    }
    \label{fig:task2}
\end{figure}

\begin{table*}[t]
    \renewcommand\arraystretch{1.1}
    \centering
    \setlength\tabcolsep{8pt} 
    \caption{A summarization of image-to-image translation tasks and methods. $\{\cdot\}$ denotes the data used in a single training. $(\cdot,\cdot)$ denotes the paired data between domains. 
    }
    \resizebox{1\linewidth}{!}{
    \begin{tabular}{lccccc}
    \hline
         \multirow{2}{*}{\textbf{Image Translation Task}} &  \multirow{2}{*}{\textbf{Representative Methods}} & \textbf{Training Data} & \textbf{Training Data} & \multirow{2}{*}{\textbf{Inference}}  \\
         &&  Fixed Domains & Adding a Domain $X$ & 
          \\
         % What scalable?
        \hline
         Paired dual-domain I2I & Pix2Pix \cite{pix2pix}, SPADE \cite{spade} & $\{(A$, $B)\}$ & - & $A \rightarrow B$   \\ \hline
         Unpaired dual-domain I2I  & CycleGAN \cite{cyclegan}, UNIT \cite{unit} & $\{A$, $B \}$ & - & $A \leftrightarrow B$  \\ \hline
         \multirow{2}{*}{Unpaired multi-domain I2I} & StarGAN \cite{stargan,starganv2}, DRIT++ \cite{drit++} & $\{A, B, C \} $ & $\{A,B,C,X\}$ & $A,B,C,X  \leftrightarrow A,B,C,X $  \\ \cline{2-5}
         & \model~(Ours)~& $\{A \}, \{B\}, \{C \} $ & $\{X \}$ & $A,B,C,X  \leftrightarrow A,B,C,X $  \\
    \hline
    \end{tabular}
    }
    \label{table:task_setting}
\end{table*}

One major weakness of existing multi-domain UNIT models is the lack of scalability, \ie, once the model is trained on certain domains, it is difficult to extend the model to unseen domains.
The reason behind this is that the shared latent space of existing methods corresponds to the domains where the training data come from. When adding new domains, the shared feature space would be changed. Therefore, one often needs to fine-tune all existing encoders and decoders to make it remap between images and the new shared latent space (Fig.~\ref{fig:task2}\red{a}). 
In this work, we present a domain-scalable UNIT method, named \emph{latent space anchoring}, which enables an improved flexibility to extend into new domains without fine-tuning on existing domains. It encodes images from each domain to the latent space of a pretrained GAN via an encoder and aligns the perceptual structure of corresponding images by the reconstruction task via a decoder. As shown in Fig. \ref{fig:task2}\red{b}, to add a new domain $X$ into our model, one only needs to train an encoder and a decoder on the data of a new domain, which are lightweight networks that embed $X$'s images to the latent space of GANs and reconstruct the images from the latent space. 
Because the GAN latent space does not depend on specific visual domains, we do not need to fine-tune all existing encoders and decoders when extending the model into new domains.
Once the domain $X$ has been incorporated into the existing UNIT model, images of the domain $X$ can be translated to any other existing domains and vice versa, since one can use the encoder of the source domain and the regressor of the target domain to translate images between any two arbitrary domains. 
The intuition and realization of our latent space anchoring method will be detailed in Sec. \ref{sec:method}.

In the experiments, we benchmark our method on various image domains (\eg, human face images, animal face images, natural images, semantic segmentation masks, sketch drawings, and landmark maps), two image-to-image translation settings (\eg, the domain-fixed UNIT setting and the domain-scalable UNIT setting), and two GANs generator backbones (\eg, StyleGAN2 \cite{stylegan2} and BigGAN-deep \cite{biggan}). 
In quantitative and qualitative comparison with the state-of-the-art methods, our method achieves impressive synthesis performances on challenging multi-domain UNIT tasks characterized by large domain gaps. This is a significant improvement over existing UNIT methods that usually only handle small domain gaps, \ie, slight changes of colors, weather conditions, or facial attributes \cite{cyclegan,stargan}. User study results also validate the superior visual quality and good structure consistency of the images produced by our method. Ablation studies on the learning objectives and network architectures further reveal the effects of different pipeline design choices. 
In summary, our work makes the following contributions: 
\begin{itemize}
    \item We propose a domain-scalable challenge of UNIT methods --- to give the models an improved flexibility to extend into new domains without training the full model on the data of all available domains.
    \item We present a novel UNIT method named latent space anchoring to tackle the  challenging task. It aligns the semantic structures of multi-domain images by  reconstructing single-domain images from the latent space of frozen GANs. 
    \item We conduct comprehensive experiments on various image domains, UNIT tasks, and different powerful GANs backbones. Qualitative and quantitative comparisons with the state-of-the-art methods, as well as the user study, validate the effectiveness of the proposed method.
\end{itemize}

\section{Related Works}

\subsection{Unsupervised Image Translation (UNIT)}
The unpaired/unsupervised image translation task aims to translate 
unpaired images between two domains~\cite{almahairi2018augmented,kim2019u,cut,zhao2020unpaired,chen2020reusing,councilgan,shao2021spatchgan,zhao2021unpaired,pizzati2021comogan}. One popular approach is based on the cyclic consistency, \eg, CycleGAN \cite{cyclegan}, DiscoGAN \cite{discogan}, and DualGAN \cite{dualgan}, which enforces synthesized images to be faithfully translated back to their original domain. CUT \cite{cut} further learns patch-wise correspondence between input and synthesized image patches via contrastive learning based on the CycleGAN \cite{cyclegan} scheme. Another approach, \eg, UNIT \cite{unit}, embeds images of different domains to one shared latent space. In this work, we use the latent space of a pretrained GAN as the intermediate anchor and independently align domains to the anchor to achieve the translation between domains.

\begin{figure*}[t]
    \centering
    \includegraphics[width=1.0\linewidth]{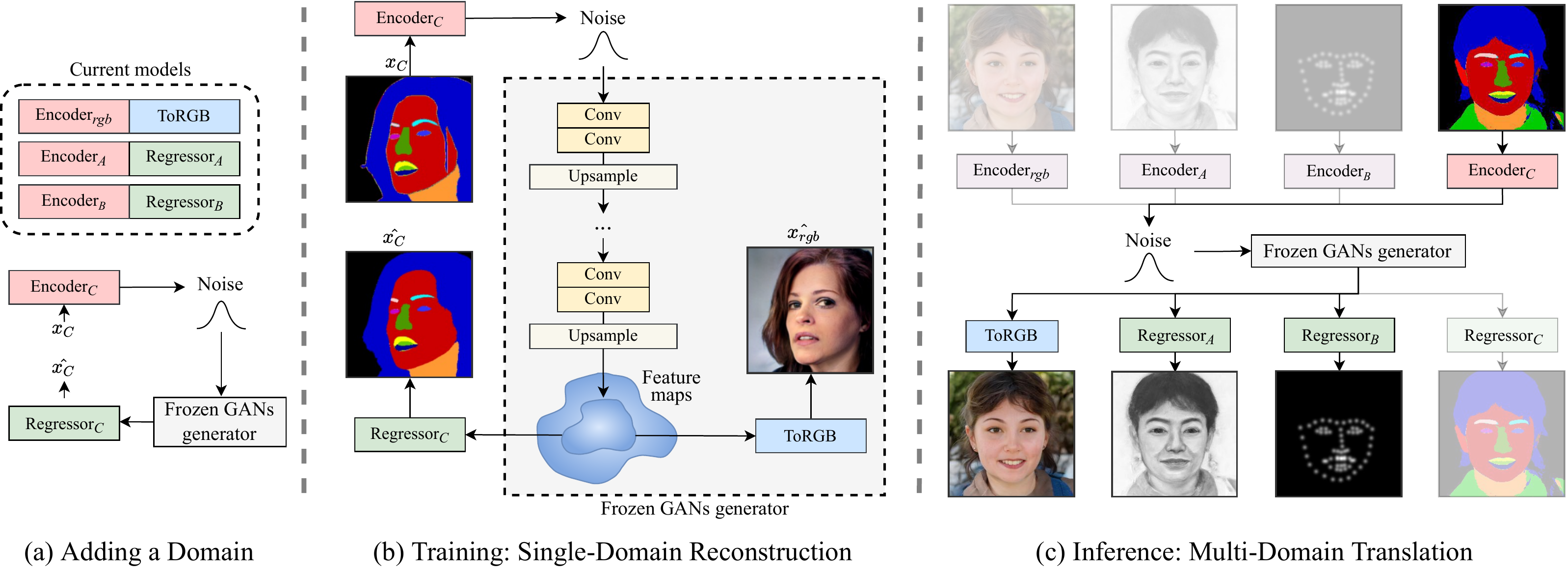}
    \caption{Our framework for domain-scalable UNIT. 
    \textbf{(a)} Adding a new domain $C$ to the current models is easy that one only needs to train an Encoder$_C$ and a Regressor$_C$ on the new domain.
    \textbf{(b)} The training is performed on the single-domain data. We use a lightweight Encoder$_C$ to map the visual domain to GAN's latent space (Sec.~\ref{sec:encode}), then use a lightweight Regressor$_C$ to reconstruct input images from GAN's latent space. We show that it is equivalent to aligning the perceptual structures of images of different domains.
    \textbf{(c)} During inference, we can use the trained encoder and regressor to perform translation between any two domains, as well as generate well-aligned multi-domain images by sampling from the GAN's latent space.
    }
    \label{fig:framework}
\end{figure*}

\subsection{UNIT of Multiple Domains} % and Multimodal UNIT}
To efficiently translate images among more than two domains, multi-domain image translation methods~\cite{liu2020gmm,xu2021domain,vinod2021multi}, \eg, StarGAN~\cite{stargan}, build unified models to learn one-to-one mappings between the shared latent space and input domains.
To sample multimodal translation results between domains, MUNIT~\cite{munit} and DRIT~\cite{drit} decompose the latent space into a shared content space and an unshared style space. Recent works~\cite{zhao2018modular,drit++,pumarola2018ganimation,funit,romero2019smit,wang2019sdit,chen2020domain,tunit,liu2021smoothing} combines the multi-domain and multimodal UNIT.
For more details, we refer readers to~\cite{pang2021image}.

Despite the progress, it remains hard to efficiently update an existing model with new domains, which often involves training or fine-tuning all previous models.
To avoid such fine-tuning, ComboGAN \cite{combogan} trains a separate encoder and decoder for the new domain and aligns its latent space to existing domains via cyclic consistency loss. 
However, it still requires images from existing domains during training.
In contrast, our proposed method only needs the new-domain images. We briefly summarize different image translation tasks and corresponding representative methods in Table \ref{table:task_setting}. Distinct to the existing image translation methods, our method is trained on images of single domains based on the latent space anchoring algorithm. In addition, it is domain-scalable, \ie, it can be efficiently scalable to unseen domains with much fewer training efforts.  

\begin{figure*}[t]
    \centering
    \includegraphics[width=1\linewidth]{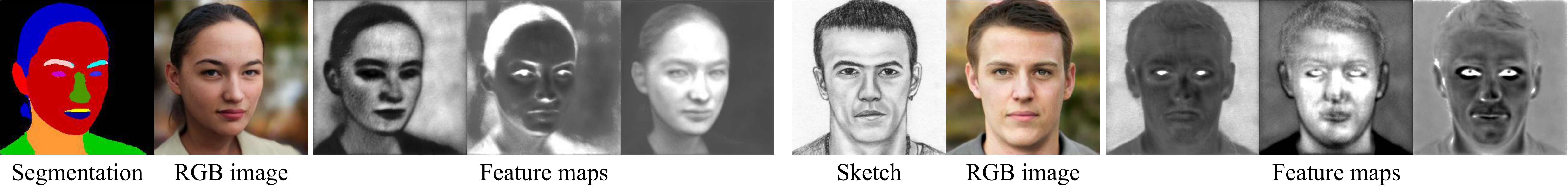}
    \caption{The visualization results of StyleGAN2 \cite{stylegan2} feature maps. We show a channel-wise visualization of randomly selected feature maps from the layer right before the ToRGB module of StyleGAN2.
    }
    \label{fig:featmap}
\end{figure*}

\subsection{GANs as Image Priors}

\noindent
\textbf{General applications.} Many existing works have employed pretrained GANs as image priors to various downstream tasks. One approach, \eg, GAN-Dissection \cite{gandissection} enables the manipulation of the semantics of images generated by GANs by discovering their relationship with feature representations.
Another approach uses GAN inversion methods \cite{xia2021gan,image2stylegan,psp} to embed images back into the latent space of pretrained GANs, leading to applications in image editing \cite{shen2020interpreting,cherepkov2021navigating,tov2021designing}, image restoration \cite{dgp,image2stylegan}, style transfer \cite{image2stylegan,image2stylegan++}, and interpreting the latent directions \cite{shen2021closed,voynov2020unsupervised}.
The proposed method is motivated by the observation that feature maps of pre-trained GAN generators contain rich and semantic structure information. A similar idea has been studied by~\cite{datasetgan} and~\cite{segmentationinstyle}, where~\cite{datasetgan} uses pre-trained GAN features for few-shot image annotation generation and~\cite{segmentationinstyle} utilizes GAN prior to perform unsupervised semantic segmentation. ~\cite{datasetgan,segmentationinstyle} can only perform pixel-aligned translation like image-to-segmentation and cannot work well on domains with structural changes. In this work, we significantly expand the application scope of this idea to domain-scalable image translation by anchoring images from different visual domains to a shared GANs prior. 

\vspace{0.5em}

\noindent{\textbf{GAN priors to UNIT.}
More relevant to our work, a few works have exploited pretrained GANs for unidirectional UNIT tasks \cite{benaim2018one,back2021fine}, \ie, GAN-to-Any UNIT. FreezeG \cite{freezeg} freezes the early layers of GANs and optimizes the other layers to generate images of target domains in adversarial training. 
It only supports mappings from the domain of pretrained GANs to other visual domains.
In addition, it does not guarantee a good perceptual structure alignment between translated images. For example, if the target domain only has frontal faces, the fine-tuned GAN can hardly generate side-view faces regardless of input images.
Instead, we propose a latent space anchoring algorithm that effectively aligns the perceptual structure of images from different domains, enabling a domain-scalable UNIT despite the limitation of the domain dataset.

\section{Method}
\label{sec:method}

\subsection{Overview}
\label{sec:framework}
In this work, we study a new challenging UNIT task, which aims at efficiently incorporating new visual domains to a learned UNIT model. The key idea of the proposed \emph{latent space anchoring} method is to use the latent space of a pretrained GAN as the {\em anchor}, where each visual domain is aligned independently during training (Fig. \ref{fig:framework}\red{a}). Then during inference, we can achieve a domain-scalable UNIT by encoding any available domain to the GAN's latent space, then decode the latent vector to any other domain without fine-tuning (Fig. \ref{fig:framework}\red{c}).
To exploit the state-of-the-art GAN models~\cite{biggan,stylegan3}, we employ an off-the-shelf generator model $G$ that maps Gaussian noise $z$ to RGB images. In general, generator $G$ can be split into the feature generator $G_\text{feat}$ and the last ToRGB layer which has only one or a few neural network layers, \ie, $G=\text{ToRGB}\circ G_{\text{feat}}$.

To align a new domain $A$ to the latent space of generator $G$, we train an encoder model $E_A$ and a regressor model $R_A$ jointly. 
We first apply $E_A$ to encode input image $x_A$ to $G$'s latent code $z_A$ to generate a visually pleasing RGB image $x_{\text{rgb}}$ (Sec.~\ref{sec:encode}):
\begin{equation} 
z_A=E_A(x_A), ~~~ f_A = G_{\text{feat}}(z_A), ~~~ x_{\text{rgb}}=\text{ToRGB}(f_A).
\end{equation}
Then, we align the perceptual structure of the input image $x_A$ and the generated $x_{\text{rgb}}$ by equivalently regressing GAN's feature maps to $\hat{x}_A$ with the regressor model $R_A$, where 
\begin{equation}
\label{eq:recon}
    \hat{x}_A =  R_A(f_A).
\end{equation}

\subsection{Image Encoding} 
\label{sec:encode}
Since there are not any matched pairs $x_A$ and $x_\text{rgb}$ in UNIT setting, learning to reconstruct $x_\text{rgb}$ from $x_A$ \cite{image2stylegan,image2stylegan++} is infeasible. Existing literature \cite{stylegan,image2stylegan,psp} shows that the distance between latent code $w$ and mean latent code $\bar{\bf{w}}$ can be used to evaluate the quality of images generated by StyleGAN \cite{stylegan}. We found that the regularization on latent code $z$ can make $z$ be close to the normal distribution $\mathcal{N}(0,\bf{I})$. 

For standard GANs, \eg~BigGAN~\cite{biggan}, we apply regularization to the encoded latent as
\begin{equation}
    \mathcal{L}_\text{latent} = \| E_A(x_A) \| _2.
    \label{eq:latent}
\end{equation}
For style-based GANs \cite{stylegan,stylegan2,stylegan3}, we encode the image into the $W^+$ latent space, and the objective function becomes $\mathcal{L}_\text{latent} = \| E_A(x_A)  - \bar{\bf{w}}\| _2$.

Besides $\mathcal{L}_\text{latent}$, the adversarial training between $\hat{x}_\text{rgb}$ and real image set $\mathcal{X}_\text{rgb}$ is optionally used to enforce $z$ to be on the latent space of pretrained GANs,
\begin{align}
    \mathcal{L}_\text{adv} = ~ & \mathbb{E}_{x_\text{rgb} \sim \mathcal{X}_\text{rgb}}[\log D_A(x_\text{rgb})]  \notag \\
   + & \ \mathbb{E}_{x_A \sim \mathcal{X}_A}[\log (1-D_A(G(E_A(x_A))))],
   \label{eq:GANloss}
\end{align}
where $D_A$ is the discriminator for domain $A$.

\label{sec:recon}

\begin{figure*}[t]
    \centering
    \includegraphics[width=1\linewidth]{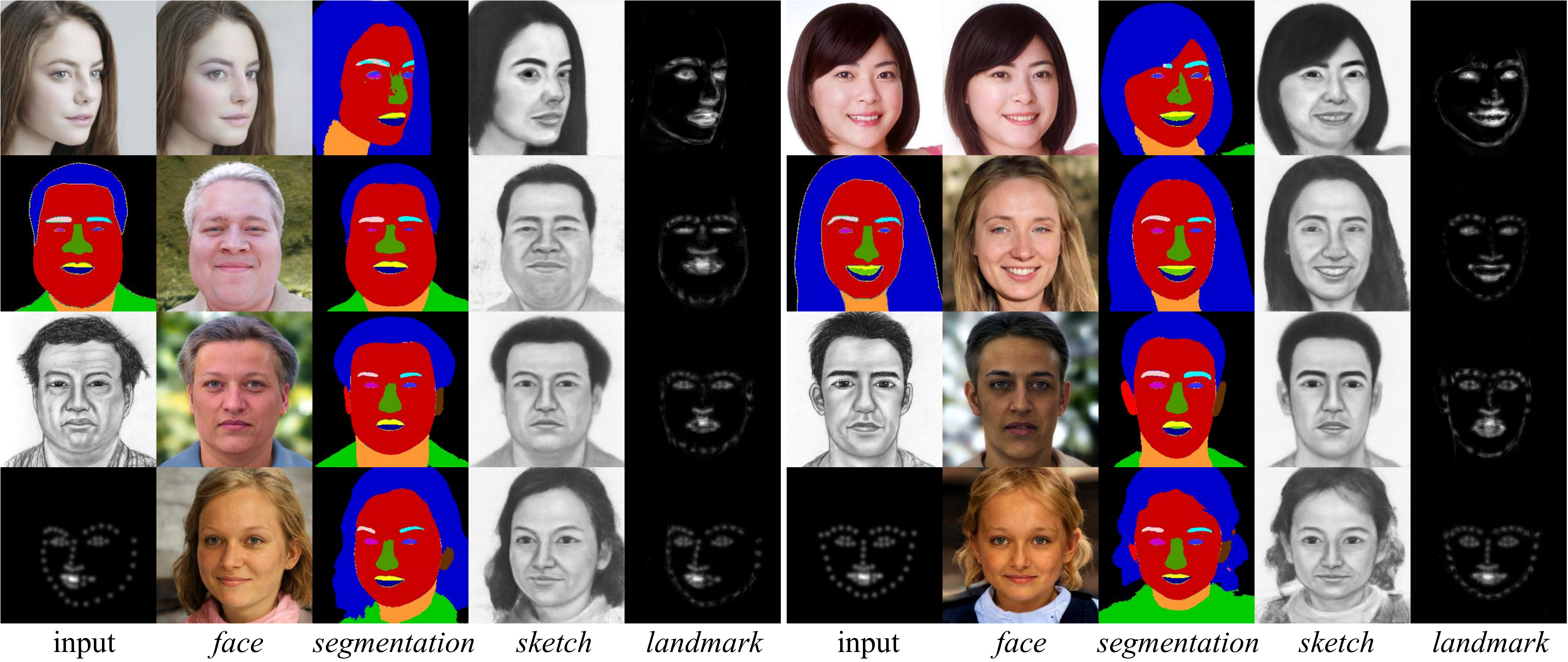}
    \caption{Unpaired image translation results of our method on four facial domains including RGB image, semantic segmentation map, sketch drawings, and landmark map. Reconstruction results are shown when inputs and outputs are of the same domain.
    }
    \label{fig:4domain}
\end{figure*}

\subsection{Latent Space Anchoring}
\noindent
\textbf{Training.} To align images from multiple domains in GAN's feature representation space, we propose a  \emph{latent space anchoring} method to align each visual domain with the latent space of a pretrained GAN, independently. The training objective for a single domain is 
\begin{equation}
    \mathcal{L}_\text{total} = \lambda_\text{rec}\mathcal{L}_\text{rec} + \lambda_\text{latent}\mathcal{L}_\text{latent} + \lambda_\text{adv}\mathcal{L}_\text{adv}.
    \label{eq:total_loss}
\end{equation}
$\lambda_\text{rec}$, $\lambda_\text{latent}$, and $\lambda_\text{adv}$ are weights of loss terms, respectively. 
$\mathcal{L}_\text{latent}$ and $\mathcal{L}_\text{adv}$ push latent features towards the latent space of the generator.
$\mathcal{L}_\text{rec}$ is the reconstruction loss\footnote{For domains such as landmark maps and sketch drawings shown in Fig. \ref{fig:framework}\red{c}, the MSE loss function is adopted as the reconstruction loss. For domains such as semantic segmentation maps, the cross-entropy loss is adopted.} which ensures the regressor $R_A$ reconstructs realistic images of domain $A$,
\begin{equation}
    \mathcal{L}_\text{rec} = \| \hat{x}_A -x_A \|_2^2.
    \label{eq:recon_loss}
\end{equation}
With Eq. \ref{eq:total_loss}, the perceptual alignment of domain $A$ and pretrained latent space is achieved by the inductive bias of models, \ie, the easiest way for models to reconstruct realistic images of domain $A$ from latent features is to project samples with the same perceptual structure into the same latent feature.
More empirical analysis of the three loss terms can be found in Fig. \ref{fig:loss_ablation}.
During training, the encoder $E_A$ learns to anchor multi-domain images to the same latent feature space and the regressor $R_A$ learns to recover multi-domain images from the latent feature space, while the pre-trained generator $G$ is fixed. 

It is of the pivotal importance to find a semantically rich latent space where images from different domains can be anchored to. Fig.~\ref{fig:featmap} shows randomly sampled feature maps right before the ToRGB layer in StyleGAN2~\cite{stylegan2}, where features of individual channels show rich, aligned, and diverse semantic structure information, \ie, the perceptual structure. Motivated by this, we use feature maps before the ToRGB layer in GAN generators as the shared anchor ground for every domain.

\noindent\textbf{Inference.}
As illustrated in Fig. \ref{fig:framework}\red{c}, the GANs latent space serves as a hub of all the domains, such that the encoders and regressors of different modalities can be arbitrarily combined to translate images from either input images (\ie, cross-domain image translation) or random noises (\ie, multi-domain image dataset generation).
Assume we have trained encoders and regressors for domain $A$ and domain $B$. When a new domain $C$ comes, we only need to train an encoder $E_C$ and a regressor $R_C$ for domain $C$, where the training process is identical to that of domain $A$ and domain $B$. The images of domain $C$ are aligned with those of domains $A$ and $B$ via the generative latent space $\mathcal{H}$, thus image translation can be performed between any two of these domains.

\section{Experiments}

\subsection{Experimental Setup}

\subsubsection{Datasets} 
In the experiments\footnote{The code of this paper is available at \url{https://github.com/siyuhuang/Latent-Space-Anchoring}}, we evaluate the proposed method on several datasets of multi-domain images. The first dataset consists of facial images from four domains: RGB images, semantic segmentation masks, sketch drawings, and landmark maps. The RGB images come from CelebAMask-HQ \cite{celebamaskhq}. The facial segmentation masks come from CelebAMask-HQ \cite{celebamaskhq}, consisting of 30,000 maps of 19 semantic classes. The sketch drawings come from CUHK Face Sketch FERET Database (CUFSF) \cite{cufsf1,cufsf2}, consisting of 1,194 facial sketch images. The facial landmarks are extracted from FFHQ \cite{stylegan} using the face pose estimator \cite{landmark} implemented in Dlib toolkit \cite{king2009dlib}. We aggregate the 68 landmarks to one channel to facilitate the training of image translation baselines. 
The second dataset consists of human and animal facial images from four domains: human facial images, dog facial images, cat facial images, and wild facial images. The animal images come from the AFHQ dataset~\cite{starganv2}, and the human faces come from the CelebAMask-HQ dataset~\cite{celebamaskhq}. 
The third dataset consists of natural images from two domains: ImageNet RGB images \cite{imagenet} and foreground object segmentation masks that are extracted by an off-the-shelf method~\cite{imagenetseg1,imagenetseg2}. 

\begin{figure*}[t]
    \centering
    \includegraphics[width=0.98\linewidth]{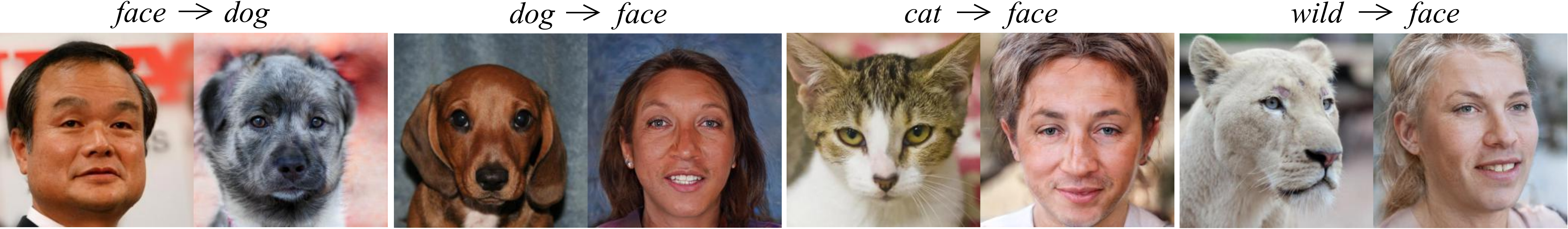}
    \caption{Unpaired image translation results delivered by our method between domains of CelebaHQ \cite{celebamaskhq}, AFHQ-dog \cite{starganv2}, AFHQ-cat \cite{starganv2}, and AFHQ-wild \cite{starganv2}.
    }
    \label{fig:face-cat-dog-wild}
\end{figure*}

\begin{figure*}[t]
    \centering
    \includegraphics[width=0.49\linewidth]{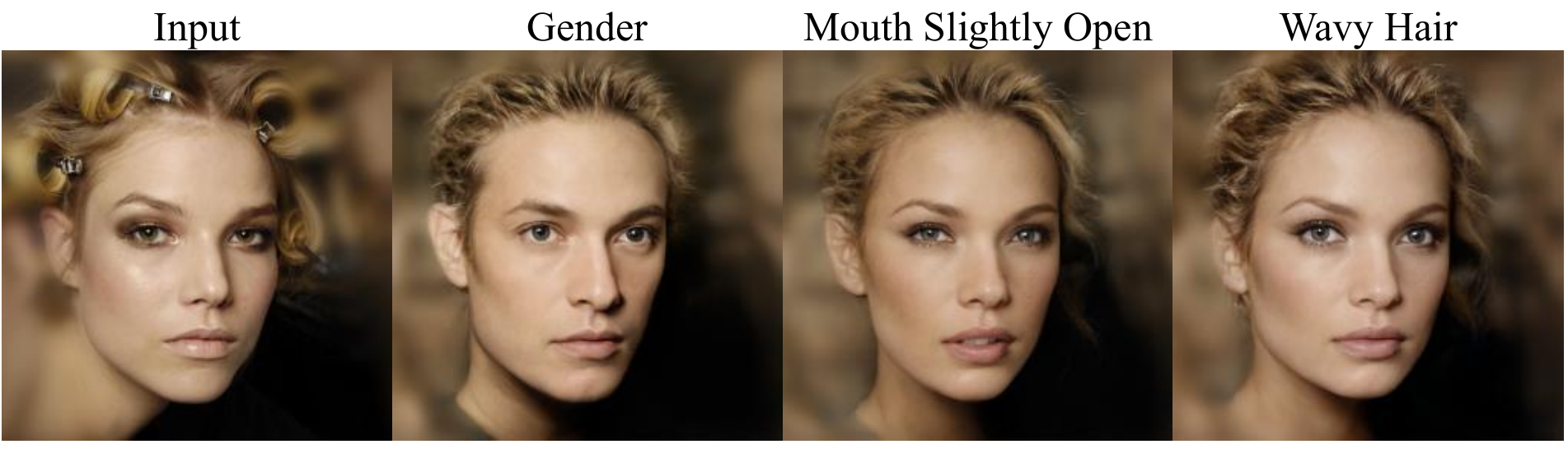}
    \includegraphics[width=0.49\linewidth]{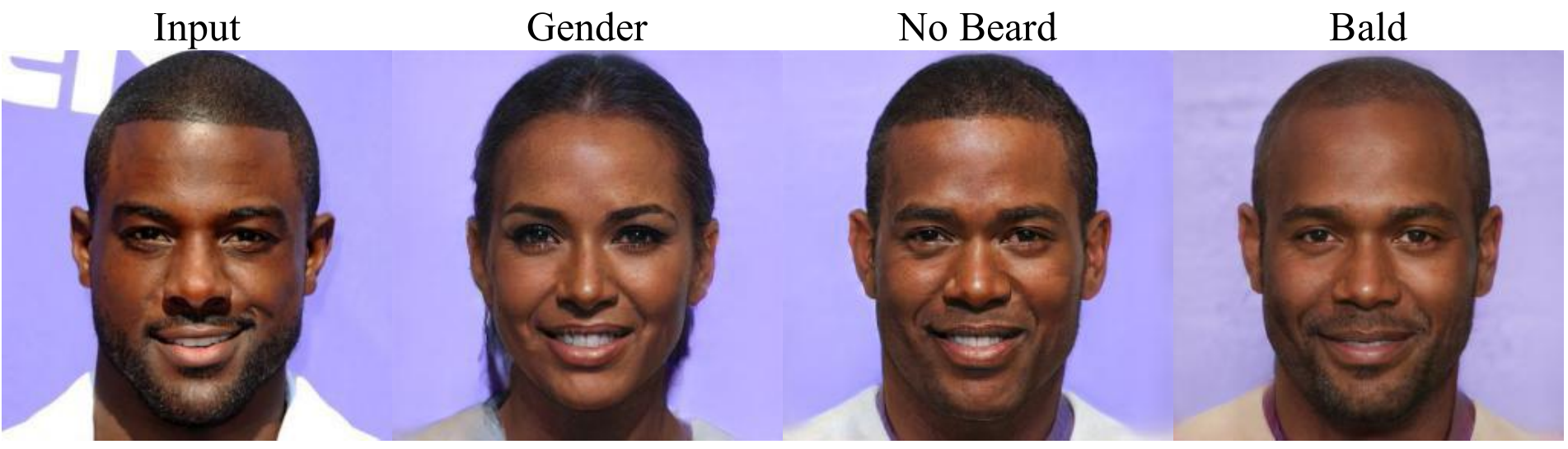}
    \caption{Unpaired facial attribute translation results of our method on the CelebAMask-HQ dataset \cite{celebamaskhq}.} 
    
    \label{fig:attribute}
\end{figure*}

\begin{figure*}[t]
    \centering
    \includegraphics[width=0.95\linewidth]{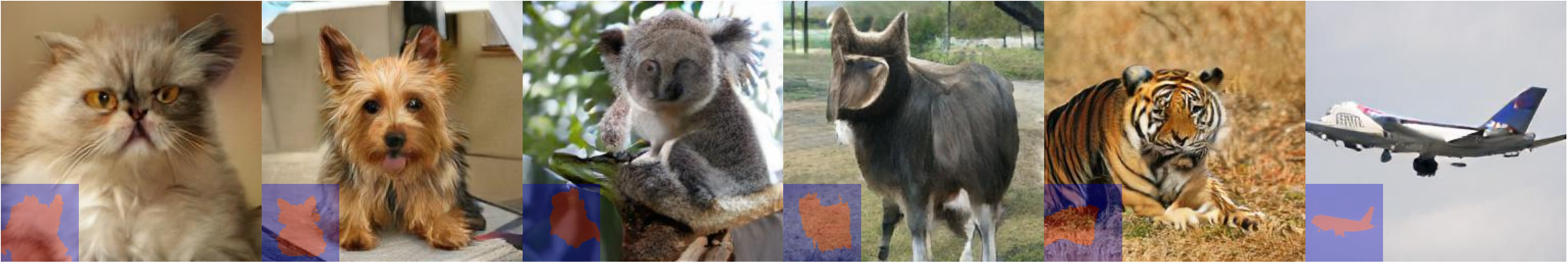}
    \caption{Unpaired image translation between foreground segmentation masks and ImageNet images. The BigGAN-deep \cite{biggan} model is employed as the generator backbone. 
    }
    \label{fig:imagenet_qual}
\end{figure*}

\subsubsection{Network and learning details} 
Our encoder network architecture follows the pSp encoder \cite{psp} which is based on a feature pyramid over the ResNet backbone \cite{resnet}. The StyleGAN2 \cite{stylegan2} and the BigGAN-deep \cite{biggan} are used as the pretrained generators for face and ImageNet image translation, respectively. The StyleGAN2 \cite{stylegan2} is pretrained on CelebAMask-HQ \cite{celebamaskhq}, while the BigGAN-deep \cite{biggan} is pretrained on ImageNet \cite{imagenet}.
We embed images into either 128-dimensional $z$ space of BigGAN-deep or 18 $W^+$ space of StyleGAN2. The regressor is a 6-layer fully convolutional network with feature map channels of (128, 64, 64, 32, 32). ReLU activation and batch normalization are adopted after every convolutional layer except the last layer.

For all experiments, our models are trained with the Adam \cite{adam} optimizer using 200k iterations, a learning rate of 1e$^{-4}$, and a batch size of 4. 
For the training objective, $\lambda_\text{latent}=0.005$, $\lambda_\text{adv}=0.01$, $\lambda_\text{rec}=1$ for the segmentation domain, and $\lambda_\text{rec}=10$ for the sketch and landmark domains.
For all translation tasks, we follow the original training/test split. Random 80\%/20\% split is adopted if the original split is not provided. All images are resized to 256$\times$256.

\subsubsection{Evaluation metrics}
We compute FID \cite{fid} to evaluate the quality of translated images. All of the paired visual domains, except sketch-to-face, have pairwise annotations. We compute LPIPS \cite{lpips} between translations and ground truth images to evaluate the correspondences learned by UNIT methods.

\begin{table*}[t]
\caption{Quantitative results for the fixed-domain UNIT setting on both facial and ImageNet images. }
\resizebox{1\linewidth}{!}{

\begin{tabular}{l|rr|c|rr||rr|rr|rr}
\hline
\multirow{3}{*}{\textbf{Method}} & \multicolumn{5}{c||}{Facial Images} &\multicolumn{6}{c}{ImageNet Images} \\\cline{2-12}
&\multicolumn{2}{c|}{\textbf{seg$\rightarrow$face}} & \textbf{sketch$\rightarrow$face} & \multicolumn{2}{c||}{\textbf{landmark$\rightarrow$face}} & \multicolumn{2}{c|}{\textbf{seg$\rightarrow$cat}} & \multicolumn{2}{c|}{\textbf{seg$\rightarrow$dog}} & \multicolumn{2}{c}{\textbf{seg$\rightarrow$plane}} \\
 & \texttt{FID}$\downarrow$ & \texttt{LPIPS}$\downarrow$ & \texttt{FID}$\downarrow$ & \texttt{FID}$\downarrow$ & \texttt{LPIPS}$\downarrow$ & \texttt{FID}$\downarrow$ & \texttt{LPIPS}$\downarrow$ & \texttt{FID}$\downarrow$ & \texttt{LPIPS}$\downarrow$ & $\texttt{FID}\downarrow$ & \texttt{LPIPS}$\downarrow$ \\ \hline
CycleGAN \cite{cyclegan} & 132.6 & 0.550 & 120.8 & 328.1 & 0.607 & 278.3 & 0.709 & 317.1 & 0.745 & 172.1 & 0.716 \\
CUT \cite{cut} & \textbf{38.9} & \textbf{0.435} & 87.6 & 287.6 & 0.595 & 342.9 & 0.699 & 350.3 & 0.752 & 148.3 & \textbf{0.695} \\
MUNIT \cite{munit} & 212.1 & 0.623 & 266.4 & 184.0 & 0.665 & 339.9 & 0.820 & 372.4 & 0.902 & 312.7 & 0.764 \\
\hline
StarGANv2 \cite{starganv2} & 53.7 & 0.532 & 85.4 & 99.5 & 0.557 & - & - & - & - & - & - \\
DRIT++ \cite{drit++} & 132.2 & 0.511 & 107.5  & 305.4 & 0.859 & - & - & - & - & - & - \\ \hline
\model~(ours) & 83.8 & 0.442 & \textbf{84.3} & \textbf{88.1} & \textbf{0.510} & \textbf{77.9} & \textbf{0.661} & \textbf{79.7} & \textbf{0.654} & \textbf{44.3} & 0.713 \\ \hline
\end{tabular}

}
\label{table:quantitative}
\end{table*}

\begin{figure*}[t]
    \centering
    \includegraphics[width=0.95\linewidth]{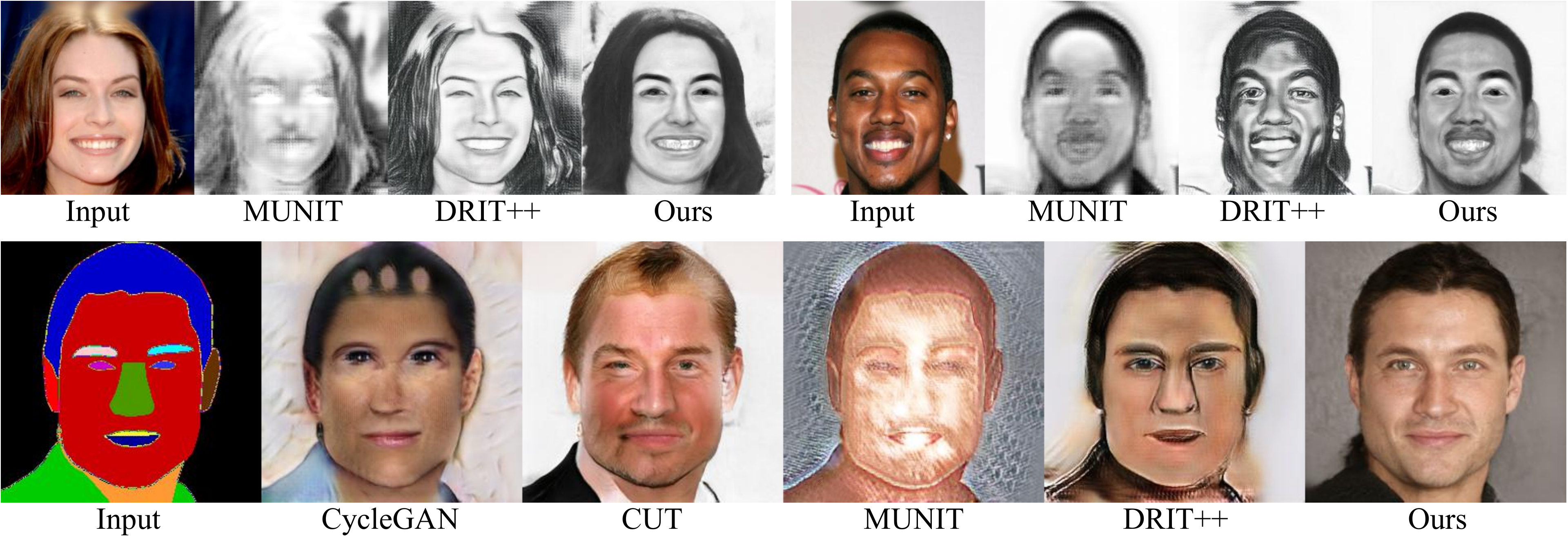}
    \caption{A comparison of the state-of-the-art unpaired image translation methods on facial image domains. \textbf{(Top)} face-to-sketch results. \textbf{(Bottom)} segmentation-to-face results.}
    \label{fig:face_qual}
\end{figure*}

\subsection{Method Results}

\subsubsection{UNIT of RGB, segmentation, sketch, and landmark}
Fig. \ref{fig:4domain} shows several examples of facial image translation by our method. The first column shows the input image, and the rest columns show the translations to four domains. We show the reconstruction results if the input and output domains are identical. Fig. \ref{fig:4domain} demonstrates that our method achieves robust translations between any two domains with diverse facial appearance details, genders, and head poses.

\subsubsection{UNIT of cat, dog, wild, and human face}
Fig.~\ref{fig:face-cat-dog-wild} shows examples of translation between domains of cat, dog, wild, and human face with a StylGAN2 backbone pretrained on the CelebAMask-HQ dataset~\cite{celebamaskhq}. 
Fig.~\ref{fig:face-cat-dog-wild} shows that our method achieves faithful AFHQ~$\leftrightarrow$~CelebaHQ translations. The shapes, face appearances, and expressions of translations well follow those of the inputs. The color of human skin is bound with the color of fur of the animal face.
In addition, the 4th example in Fig.~\ref{fig:face-cat-dog-wild} shows that the head pose (\eg, orientation) of translation is consistent with the pose of the input.

\subsubsection{UNIT of facial attributes}
Fig. \ref{fig:attribute} shows the results of our method for facial attribute translation. 
We select five different attributes from CelebAMask-HQ \cite{celebamaskhq}, including Gender, Mouth Slightly Open, Wavy Hair, No Beard, and Bald, as they cover human facial information from global structures to local details. The attribute translation task is formulated as a multi-domain UNIT task, where each domain corresponds to a positive/negative attribute. 
As shown in Fig. \ref{fig:attribute}, our method generates vivid images and changes the attributes of input images accurately. It demonstrates that our method can work on UNIT tasks with some degree of perceptual structure change.

\subsubsection{UNIT of mask and natural image}

In Fig. \ref{fig:imagenet_qual}, we show segmentation-to-RGB translation results of our method on six different ImageNet classes \cite{imagenet}. Thanks to the BigGAN-deep generator backbone \cite{biggan}, the synthesized images are highly realistic and distinct. The synthesized images faithfully follow the shapes provided in segmentation masks. Although the input segmentation masks are imperfect, \eg, the right ear of the cat is missing, our method learns a robust mapping between segmentation masks and RGB images to restore the unreasonable parts of masks.

\subsection{Comparison with State-of-the-art Methods}
Here, we compare the proposed method with the state-of-the-art methods on two UNIT settings: standard multi-domain UNIT (see Sec. \ref{sec:standardUNIT}) and domain-scalable UNIT (see Sec. \ref{sec:domainscalableUNIT}). We also compare our method with three pretrained GANs-based UNIT methods in Sec. \ref{sec:GANsUNIT}.

\subsubsection{Standard multi-domain UNIT setting}
\label{sec:standardUNIT}

\noindent\textbf{Baseline methods.}
We compare five UNIT methods, including three representative dual-domain methods, \ie, CycleGAN \cite{cyclegan}, CUT \cite{cut}, and MUNIT \cite{munit}, and two representative multi-domain methods, \ie, StarGANv2 \cite{starganv2} and DRIT++ \cite{drit++}.

\noindent\textbf{Quantitative results.}
Table \ref{table:quantitative} shows that our method achieves the best FID and LPIPS in most cases. CUT \cite{cut} shows the best performance for segmentation-to-face. We conjecture it is because the outputs of CUT \cite{cut} does not strictly match the input segmentation map and thus have a higher diversity. The results demonstrate the effectiveness of our method for standard UNIT tasks.

\begin{figure}[t]
    \centering
    \includegraphics[width=1\linewidth]{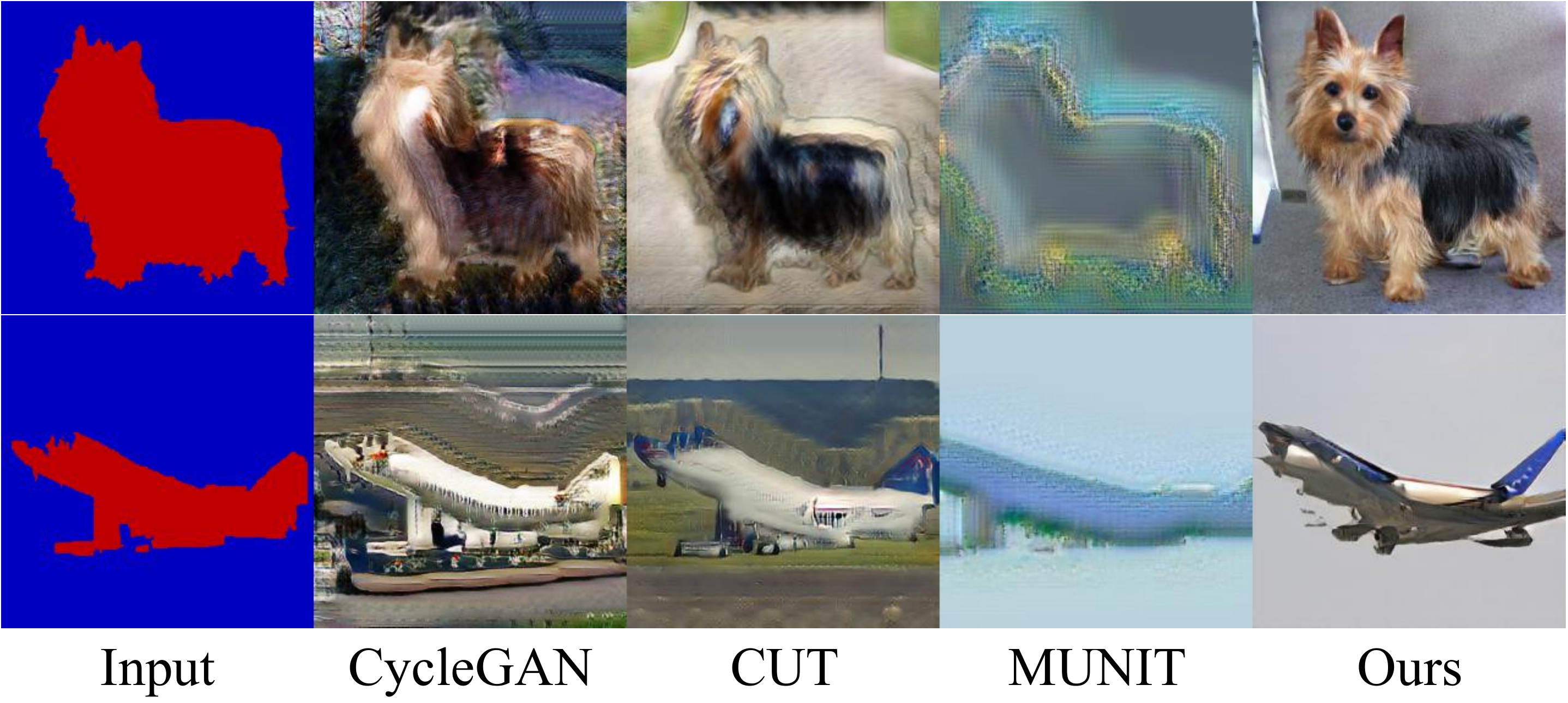}
    \caption{A comparison of the unpaired image translation methods for translating segmentation masks to ImageNet images \cite{imagenet}.
    }
    \label{fig:imagenet_comparison}
\end{figure}

\begin{figure}[t]
    \centering
    \includegraphics[width=0.81\linewidth]{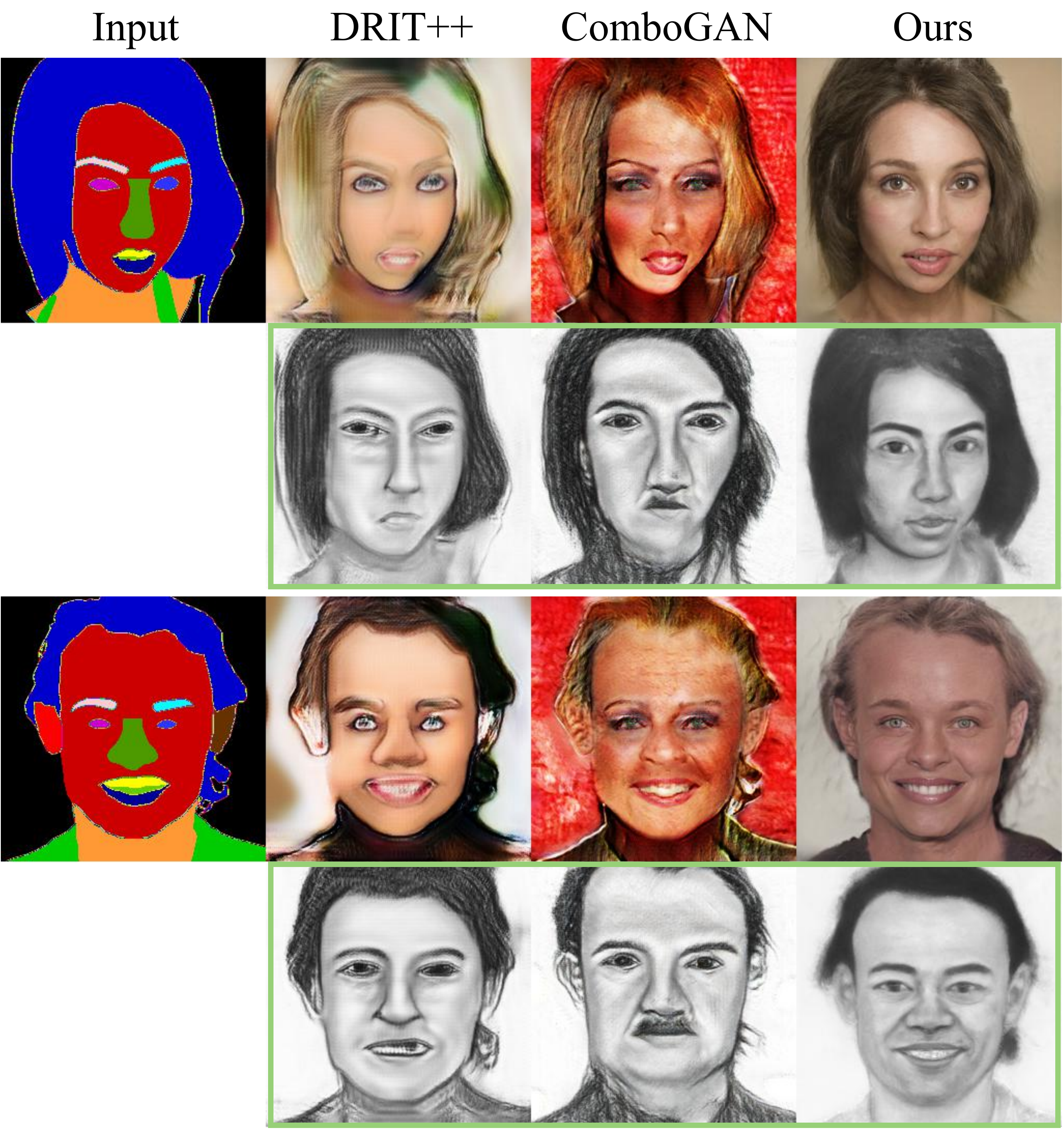}
    \caption{Results of the domain-scalable UNIT task, which aims at adding a new sketch drawing domain (as denoted in \textcolor{green}{green} bounding boxes) to existing UNIT models. 
    }
    \label{fig:adddomain}
\end{figure}

\begin{figure*}[t]
    \centering
    \includegraphics[width=1\linewidth]{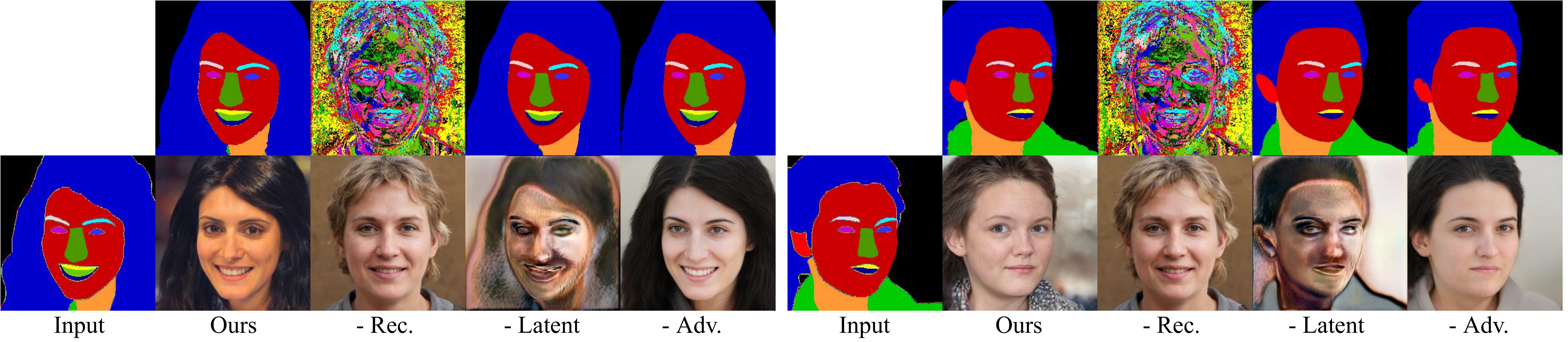}
    \caption{An ablation study of our training objectives by removing one loss at a time. We show the \textbf{(top)} reconstructed segmentation masks and the \textbf{(bottom)} generated RGB images. ``- Rec'': removing the reconstruction loss in Eq. \ref{eq:recon_loss}. ``- Latent'': removing the latent regularization loss in Eq. \ref{eq:latent}. ``- Adv.'': removing the adversarial loss in Eq. \ref{eq:GANloss}. 
    }
    \label{fig:loss_ablation}
\end{figure*}

\begin{figure}[t]
    \centering
    \includegraphics[width=1\linewidth]{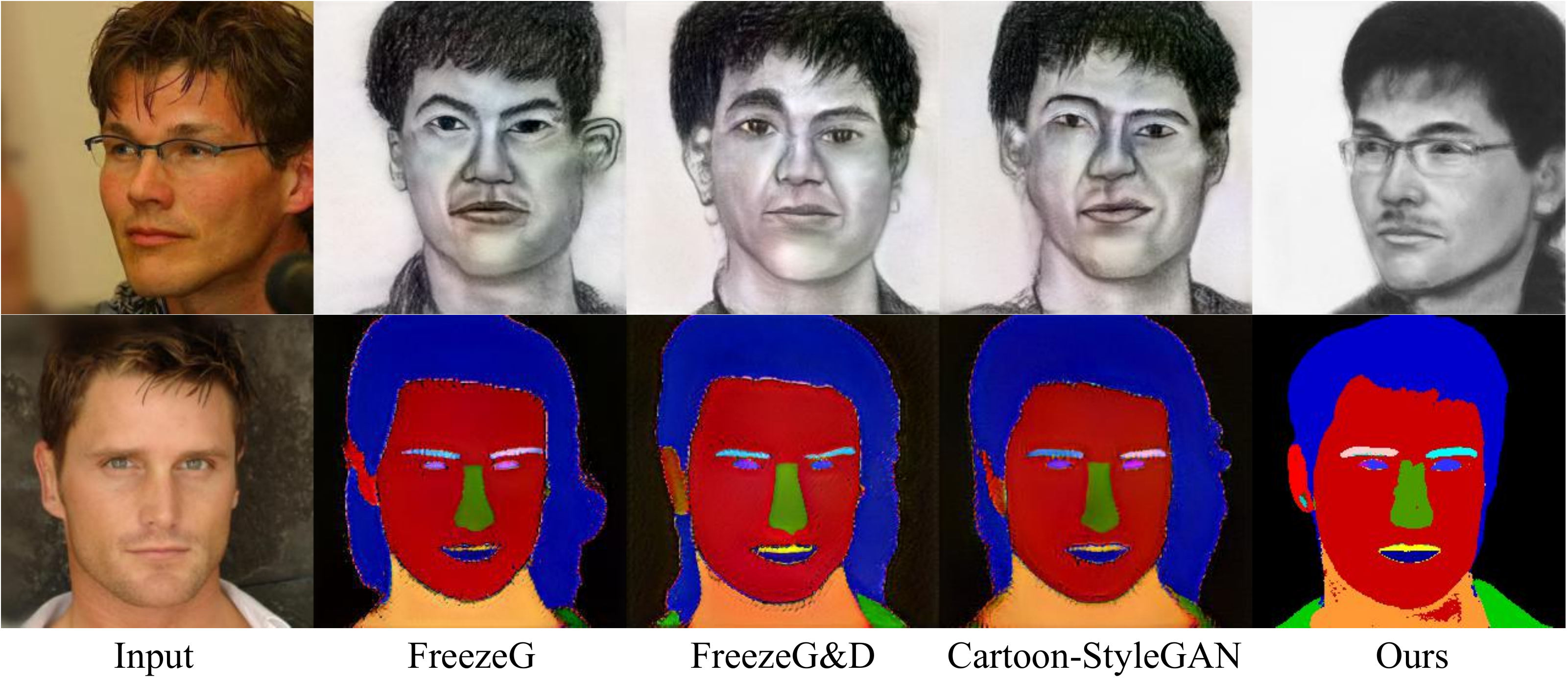}
    \caption{A comparison of pretrained GANs-based unpaired image translation methods for RGB-to-sketch and RGB-to-segmentation translations. 
    }
    \label{fig:freezeg}
\end{figure}

\noindent\textbf{Qualitative results.}
Fig. \ref{fig:face_qual} shows a qualitative comparison of the baseline methods for translating between multiple domains of facial images. The first row shows the results of translating an RGB image to a sketch (drawing). Both DRIT++ \cite{drit++} and our method show reasonable results. DRIT++ is basically built upon feature style transfer, thus its synthesis images look sharper but contain more visual artifacts, as shown in the first example of Fig. \ref{fig:face_qual}. The second row shows the results of translating semantic segmentation to RGB images. Our method delivers the most vivid and realistic translation result. CUT \cite{cut} is the second-best method for segmentation to RGB images due to its effective contrastive learning framework. 

In Fig. \ref{fig:imagenet_comparison}, we qualitatively compare our method with the dual-domain UNIT methods for translating foreground segmentation masks to ImageNet images. Our method delivers vivid dog and plane images. The baseline methods show poorer performances on this task, mainly because there are fewer training samples (\ie, 1000 images per domain) in this experiment, compared against the segmentation-to-face experiment (27k images per domain). 

\subsubsection{Domain-scalable UNIT setting}
\label{sec:domainscalableUNIT}
The task is to extend an off-the-shelf UNIT model (\eg, trained for segmentation maps and RGB images) to a new visual domain (\eg, sketch drawings) provided a set of images of the new domain.

\noindent\textbf{Baseline methods.} We compare our method with a multi-domain method DRIT++ \cite{drit++} and a unrestricted-domain method ComboGAN \cite{combogan}. For the pretrained dual-domain DRIT++~\cite{drit++} model, we fine-tune it with the data of all three available domains. For the pretrained dual-domain ComboGAN~\cite{combogan} model, we add an encoder and a decoder that optimize the cyclic consistency loss between the sketch domain and both two existing domains.

\noindent\textbf{Qualitative results.} 
Fig. \ref{fig:adddomain} shows qualitative results of the  domain-scalable UNIT setting. First, the proposed method generates more visually-pleasing facial images compared with baseline methods even on existing domains, \ie, segmentation-to-RGB, thanks to the powerful pretrained GANs generator.
Moreover, for the newly added sketch domain, sketch images translated by the proposed method are better aligned to the input segmentation masks than the baseline methods. The two baseline methods cannot well align the perceptual structure of the new domain to those of previous domains. The results validate the superior scalability of our method for new visual domains.

\begin{figure*}[t]
    \includegraphics[width=1\linewidth]{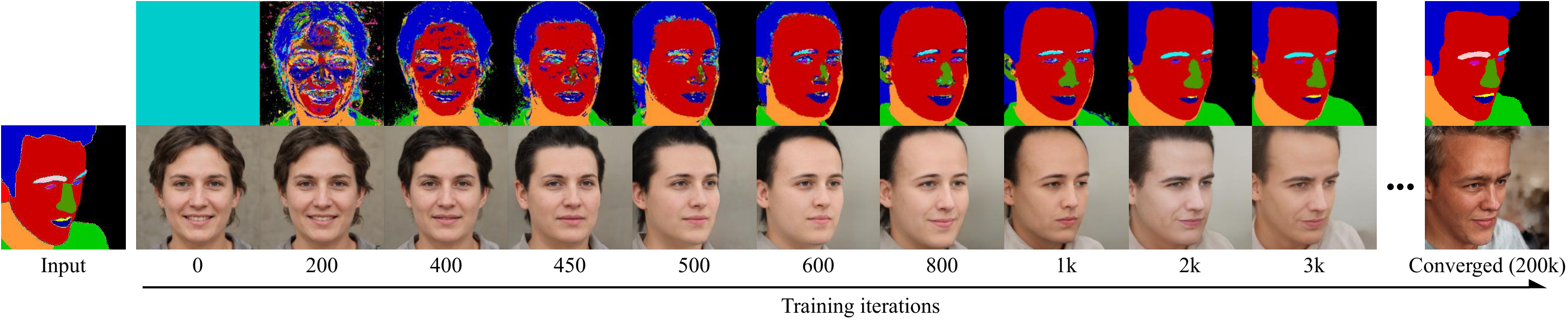}
    \caption{The evolution of \textbf{(top)} reconstructed masks and \textbf{(bottom)} generated RGB images during training.}
    \label{fig:training_process}
\end{figure*}

\subsubsection{Pretrained GANs-based UNIT methods} 
\label{sec:GANsUNIT}

\noindent\textbf{Baseline methods.} FreezeG \cite{freezeg} and Cartoon-StyleGAN \cite{freezesg} are pretrained GANs-based UNIT methods. To extend to new visual domains, they freeze the lower layers of GANs generators while fine-tuning the other layers. To make them be able to encode images into GAN's latent space, we adopt an effective optimization-based GAN inversion method~\cite{inversion}.

\noindent\textbf{Qualitative results.} Fig. \ref{fig:freezeg} shows that the two baseline methods fail to achieve a good visual alignment of inputs and predictions. For the face-to-sketch translation, all training sketche drawings are of frontal faces in the CUFSF dataset \cite{cufsf1}, such that the adversarial learning scheme adopted by the baseline methods limits them to generate sketches of frontal faces only. However, our method is based on feature reconstruction that the perceptual structures of generated images well follow those of the inputs. For face-to-segmentation UNIT, the hair segmentation maps predicted by the baseline methods are not accurate enough that the hairs are mistakenly recognized as shadows around faces. Our method predicts a more accurate segmentation map. It demonstrates the effectiveness of our latent space anchoring algorithm for aligning visual domains via the latent space of pretrained GANs.

\begin{figure}[t]
    \centering
    \includegraphics[width=1\linewidth]{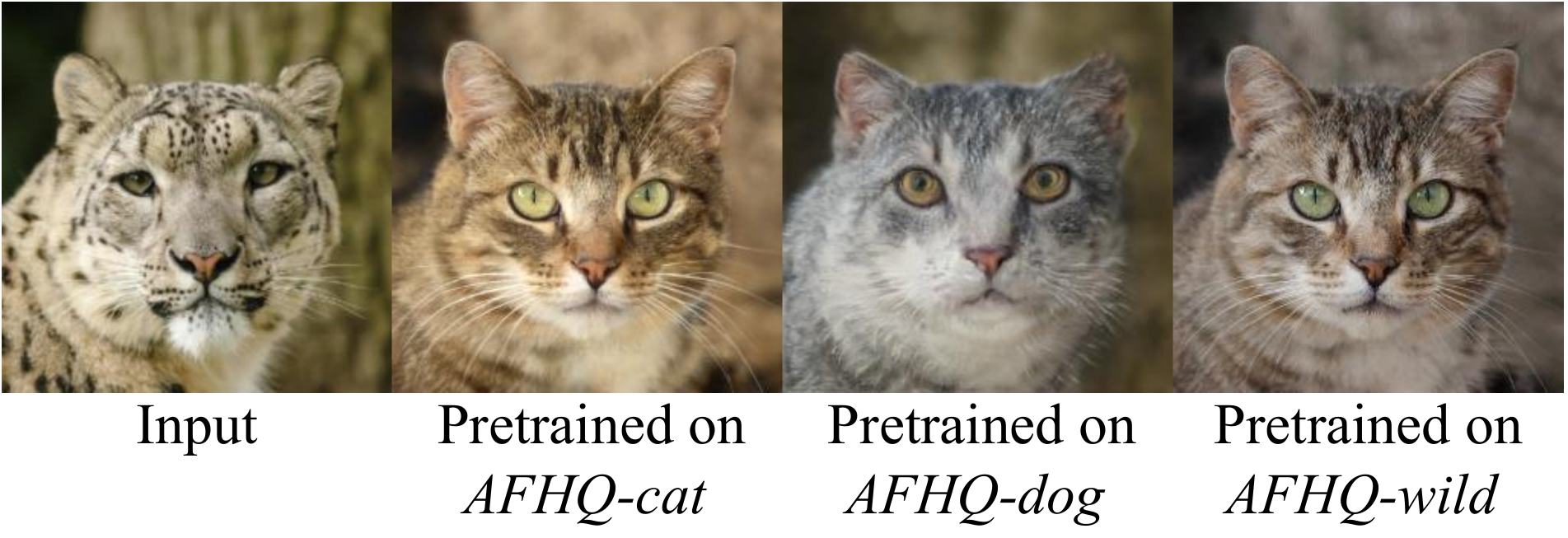}
    \caption{A comparison of generator backbones pretrained on different datasets for unpaired image translation from \emph{AFHQ-wild} to \emph{AFHQ-cat} \cite{starganv2}.}

    \label{fig:backbone}
\end{figure}

\begin{table*}[t]
\centering
\caption{Ablation studies of the input feature map size and network architecture for the regressor. Results are reported on segmentation-to-RGB task with a StyleGAN2 generator backbone \cite{stylegan2} pretrained on FFHQ dataset \cite{stylegan}. The ``All'' in the column of Feature size denotes all the feature maps from 4$\times$4 to 256$\times$256 are concatenated as the input of regressor.}
\resizebox{0.75\linewidth}{!}{
\begin{tabular}{l|c|c|cccc}
\hline
\textbf{Method} & \textbf{Feature size} & \textbf{Regressor} & \texttt{FID}$\downarrow$ & \texttt{KID}$\scriptstyle \times 10^3$ $\downarrow$  & \texttt{SSIM}$\uparrow$ & \texttt{LPIPS}$\downarrow$ \\ \hline
\model~(ours) & 256$\times$256 & 6 conv layers & \textbf{83.8} & \textbf{72.8} & \textbf{0.307} & \textbf{0.442} \\ \hline
\multirow{5}{*}{Ablation of feature size} & 4$\times$4 & - & 201.5 & 199.0 & 0.207 & 0.525 \\
 & 16$\times$16 & - & 188.9 & 217.7 & 0.269 & 0.476 \\
 & 64$\times$64 & - & 115.6 & 111.5 & 0.260 & 0.493 \\
 & 128$\times$128 & - & 116.2 & 116.2 & 0.267 & 0.481 \\ 
 & All & - & 129.5 & 129.8 & 0.279 & 0.484 \\ \hline
{\multirow{2}{*}{Ablation of regressor}} & - & 3 conv layers & 93.8 & 83.5 & 0.297 & 0.451 \\
 & - & 6 ResBlocks & 155.1 & 151.0 & 0.249 & 0.579 \\ \hline
\end{tabular}
}
\label{table:ablation}
\end{table*}

\subsection{Ablation Study}
We further conduct various ablation experiments to examine the effects of various pipeline design choices.

\subsubsection{Learning objectives} 
Fig.~\ref{fig:loss_ablation} studies the effects of the three loss terms in our training objectives, including the reconstruction loss, latent regularization loss, and adversarial loss, by separately removing one of them during training. Fig.~\ref{fig:loss_ablation} shows that the reconstruction loss is vital for our framework, as removing it results in a fail reconstruction of both images and segmentation masks. It validates that the reconstruction loss plays an key role in aligning the structures of images from different domains. Removing the latent regularization loss results in the distortion of generated images, because the encoder fails to embed input images into the latent space of GANs. The adversarial loss enhances the fidelity of the generated images, making them contain richer colors, textures, and background information. The results shown in Fig.~\ref{fig:loss_ablation} are in line with our intuition for devising the three learning objectives. 

\subsubsection{Training iteration} Fig.~\ref{fig:training_process} visualizes the evolution of model results during training. Within the first 400 training iterations, the generated images are similar to that of iteration-0, \ie, the initialized model. Beginning from iteration-400, the perceptual structures of reconstructed segmentation masks and generated images change to the input segmentation gradually. At iteration-3k, the reconstructed segmentation is basically identical to the input segmentation. Comparing iteration-3k and iteration-200k, the reconstructed segmentation masks are similar, while the image of iteration-200k shows richer appearance, color, and background details. It denotes that the model can learn latent space anchoring quickly, then improves latent space encoding afterwards.

\subsubsection{Pretrained GANs Prior}
We study the effect of generator backbones pretrained on different datasets for the translation from AFHQ-wild to AFHQ-cat \cite{starganv2}. As shown in Fig. \ref{fig:backbone}, different pretrained generator priors lead to different visual results, while all three GAN priors deliver decent image translation results, indicating the robustness of our method. In addition, when the GAN generator is pretrained on either the input or output domain, the results would contain slightly more fine-grained details. For example, the AFHQ-cat and AFHQ-wild results in Fig.~\ref{fig:backbone} show more texture details on the cat faces.

\subsubsection{Input feature maps of Regressor} 
We further study which feature maps of a StyleGAN2 \cite{stylegan2} are more effective for the proposed method. Table \ref{table:ablation} shows that the translation performances are better with higher-level feature maps, because the low-level feature maps contain fewer perceptual structures, leading to a worse latent space anchoring effect. Concatenating all the generator feature maps from 4$\times$4 to 256$\times$256 also achieves a poor performance. We conjecture it is because there are too many feature channels, which would lead to an over-strong regressor that fails the latent space anchoring process. 

\subsubsection{Network architecture of Regressor}
As discussed above, the latent space anchoring method limits the representation ability of regressors. However, a very weak regressor would not well decode the feature maps to multi-domain images. The bottom of Table \ref{table:ablation} studies different regressor network architectures, including 3 conv layers, 6 conv layers, and 6 ResNet blocks \cite{resnet}. The 6 conv layers perform the best. 3 conv layers perform slightly worse than 6 conv layers. 6 ResBlocks are much more powerful than the aforementioned two networks, but they show a poor performance since they corrupt the weak regressor assumption.

\begin{table*}[t]
\centering
\caption{A user study on \textit{segmentation map $\to$ face} task. ``Visual Quality Votes'': the percentage of a method that obtains the best visual quality votes. ``Structure Match Votes'': the percentage of a method that obtains the best structure match votes.}
\resizebox{0.92\linewidth}{!}{
\begin{tabular}{l|cccccc}
\hline
    Method & ComboGAN~\cite{combogan} & CUT \cite{cut} & CycleGAN \cite{cyclegan} & DRIT++ \cite{drit++} & MUNIT \cite{munit} & \model~(ours) \\
    \hline
    Visual Quality Votes (\%) & 1.67 & 10.56 & 0.56 & 4.72 & 1.11 & \textbf{81.39} \\
    Structure Match Votes (\%) & 4.44 & 27.22 & 4.44 & 16.39 & 7.78 & \textbf{39.72} \\
    \hline
\end{tabular}
}
\label{table:user_study}
\end{table*}

\begin{figure*}[t]
    \centering
    \vspace{1em}
    \includegraphics[width=0.85\linewidth]{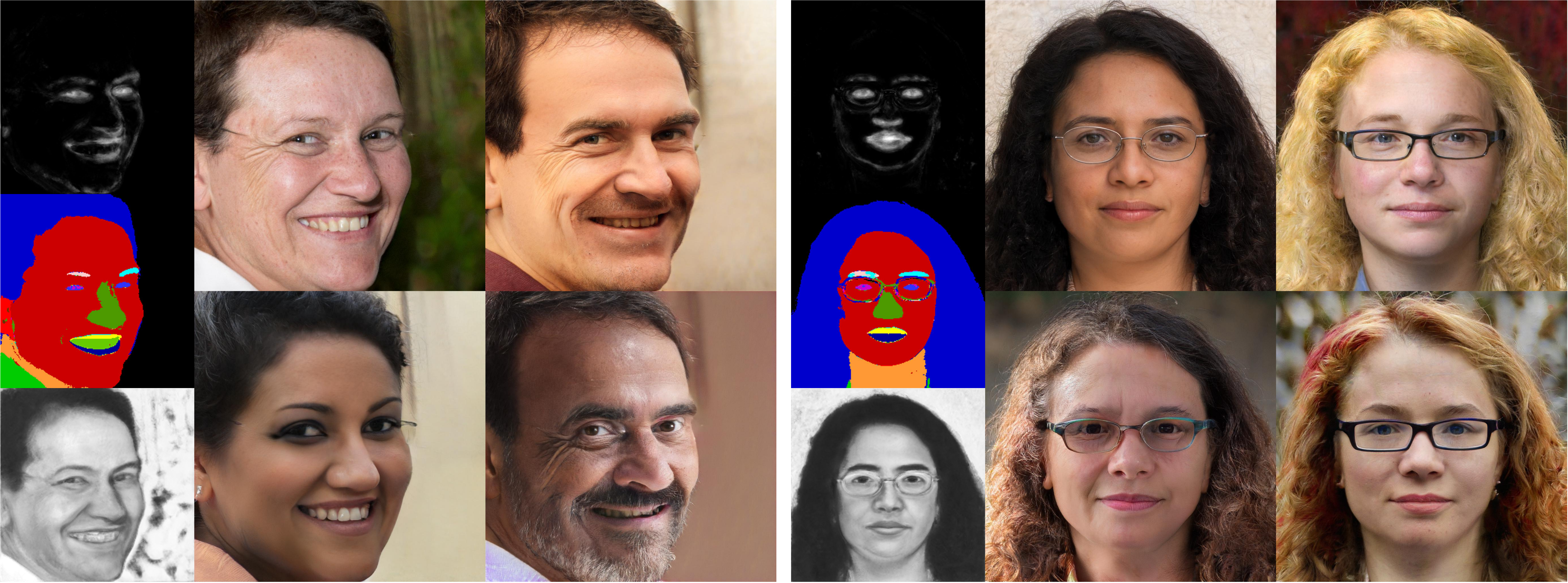}
    \caption{Generating high-resolution high-fidelity multi-domain and multimodal images via sampling from the latent space of StyleGAN2-FFHQ \cite{stylegan2}. In each example, upper left shows the image generated by originally encoded $W^+$. The rest three RGB images are the corresponding multimodal generations obtained by replacing the 5-th to 18-th (left) and 7-th to 18-th (right)
    $W^+$ of StyleGAN2 \cite{stylegan2} with random variables.}
    \label{fig:multimodal}
\end{figure*}

\begin{figure*}[t]
    \centering
    \includegraphics[width=1\linewidth]{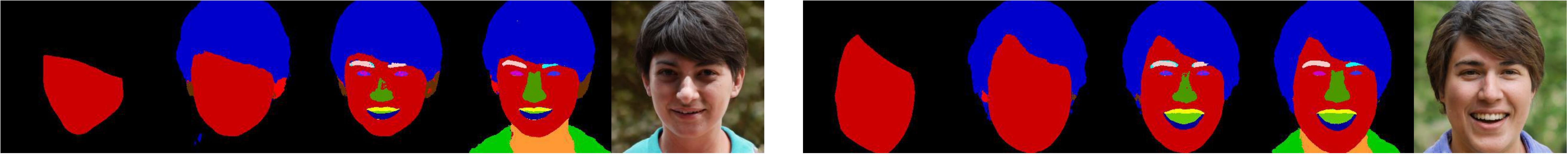}
    \caption{A progressive synthesis of human faces from coarse-grained to fine-grained semantic segmentation masks. Our method can translate the multi-domain inputs easily via anchoring them to the latent space of GANs.}
    \label{fig:progressive}
\end{figure*}

\subsection{User Study}
We conduct a user study to subjectively compare the performance of the proposed method with the state-of-the-art methods. We choose the \textit{segmentation map $\to$ face} UNIT task, and randomly collect 300 sets of samples from the image translation results to form a sample pool. For each user, we randomly select 15 sets of samples from the pool, where each set consists of an input image and the translated images from six comparison methods (CycleGAN \cite{cyclegan}, CUT \cite{cut}, MUNIT \cite{munit}, DRIT++ \cite{drit++}, ComboGAN~\cite{combogan}, and our method). For each set, we let the user choose one image of the best visual quality and one image that can best match the structure of the input. To ensure a fair user study, we design a website and place all the results of different algorithms one-by-one, where the order of images is shuffled individually for each set of results. We collected 24 questionnaires with 360 votes in total. Table \ref{table:user_study} shows the result of our user study. Our method obtains 81.39\% votes regarding the best visual quality and 39.72\% votes regarding the best structure match, both ranking the first among all the comparison methods. It demonstrates that our method can deliver images of good quality and preserve the structure of input images in terms of human evaluation.

\subsection{Additional Applications}

\subsubsection{Multimodal multi-domain dataset synthesis} 
Since our method aligns multi-domain images to the latent space of pretrained powerful GANs, we can generate infinite well-aligned, high-resolution, and highly realistic multimodal multi-domain images by sampling from the latent space of generators. As shown in Fig. \ref{fig:multimodal}, the landmark maps, semantic segmentation maps, sketch drawings are generated along with the RGB images (the upper left ones). In addition to multi-domain generation, our method also supports generating \emph{multimodal} images (\ie, images with different appearance details as presented in MUNIT \cite{munit} and DRIT \cite{drit}) by replacing the $W^+$ latent variables of StyleGAN2 \cite{stylegan2} with randomly sampled variables. Fig. \ref{fig:multimodal} shows the semantic structures of multi-modal RGB images are highly aligned with the multi-domain generations. The multi-domain and multimodal image generation would benefit many downstream visual analysis tasks, such as generating a large number of data samples to augment existing multi-domain image datasets. 

\subsubsection{Progressive image translation} Different users may prefer using inputs of various annotation levels for image translation. Here we show a progressive image translation example in Fig. \ref{fig:progressive}, where the segmentation masks vary from coarse-grained to fine-grained annotations. We train encoders and regressors for these four types of segmentation inputs individually. Thanks to latent space anchoring, we unify multiple image domains easily by anchoring new visual domains to the latent space of GANs. In inference, more fine-grained segmentation masks are inferred based on previous ones, providing a series of segmentation maps of different annotation levels to users. 

\section{Conclusion}

In this paper, we address a new challenge of unpaired image translation, \ie, domain-scalable UNIT, and presented a novel latent space anchoring method to address this new challenge. Our method aligns different visual domains to the latent space of pretrained GANs generator via learning to reconstruct images on single domains. We have comprehensively evaluated our method on diverse domains including human faces, animal faces, and ImageNet images with StyleGAN2 \cite{stylegan2} and BigGAN-deep \cite{biggan} as the  generator backbones. Quantitative, qualitative, and user study results have validated that our method achieves superior performance compared with previous UNIT methods for standard and domain-scalable UNIT tasks.

\section{Limitation and Ethical Statement}

\noindent\textbf{Limitation.} One limitation of our method is that it is devised for aligning perceptual structures of images. Therefore, only visual domains having similar perceptual structures can be incorporated into this framework. For instance, human facial images, human facial segmentation masks, and animal facial images can be aligned. However, human facial images and ImageNet foreground segmentation masks would not be well aligned as they have different perceptual structures in datasets. In the future, we plan to resolve this limitation by incorporating deformable models \cite{jaderberg2015spatial,dai2017deformable}, semantic layouts \cite{zheng2022semantic,zheng2022cm}, or geometric priors \cite{tan2020crossnet++,zhang2020cross}. 

\vspace{0.5em}

\noindent{\textbf{Ethical Statement.} The proposed method is a general multi-domain image translation framework. As shown by the experimental results, our method can be used in human and animal face translation, which is considered a potential ethical issue. To prevent the abusive usage of the proposed method, the codes and model checkpoints of the human/animation translation are not released to the public. They are only released to researchers upon formal requests and can only be used for research purposes.

\ifCLASSOPTIONcompsoc
  \section*{Acknowledgments}
\else
  \section*{Acknowledgment}
\fi
This work was partially supported by NIH grants R01HD104969 and 5U54CA225088-03.

\appendix

\noindent\textbf{An intuitive explanation of latent space anchoring.} 
To align images from multiple domains in GAN's feature representation space, we propose the \emph{latent space anchoring} method. Here we additionally provide an intuitive explanation of this method, \ie, why the single-domain anchoring results in an alignment of multiple domains in the latent space.

First, it is of the pivotal importance to find a semantically rich latent space where images from different domains can be anchored to. Fig.~\ref{fig:featmap} shows randomly sampled feature maps right before the ToRGB layer in StyleGAN2~\cite{stylegan2}, where features of individual channels show rich, aligned, and diverse semantic structure information, \ie, the perceptual structure. Motivated by this, we adopt feature maps before the ToRGB layer in GAN generators as the shared anchor ground for every domain.

\begin{figure}[t]
    \centering
    \includegraphics[width=0.35\textwidth]{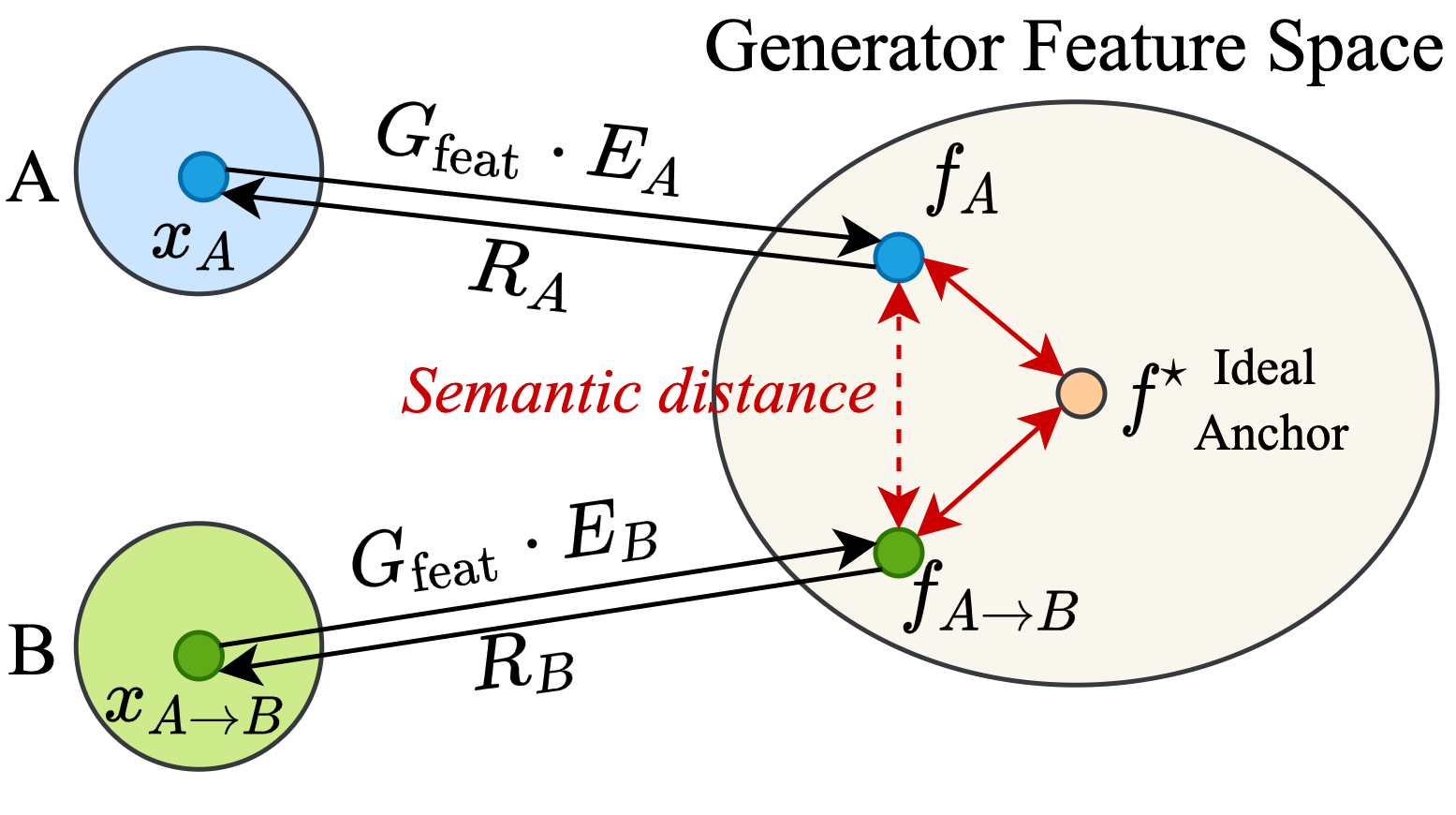}
    \caption{
    An illustration of {\em latent space anchoring}. 
    We minimize two uni-domain anchoring distances to minimize the perceptual structure gap of
    an input $x_A$ in domain $A$ and its translation $x_{A\to B}$ in domain $B$. Two anchoring distances upper bound the semantic distance between input feature $f_A$ and translated feature $f_{A\to B}$ based on the triangle inequality.}
    % \blue{remove $E_A,G_{feat}$ in (b) as they are not involved/make it simpler to see}
    % Siyu: $E_A,G_{feat}$ are involved as they are the encoders of reconstruction process.
    \label{fig:pa}
\end{figure}

Based on the pre-defined generator feature space as the anchor ground, our goal is to minimize the semantic distance between the input and the translated image, where the semantic distance is defined by the feature distance in the shared latent space. Ideally, these two images should have the same semantics but the appearance should belong to different domains. However, one cannot directly minimize the semantic distance as there is no paired training data in UNIT tasks. 

Instead, we minimize two uni-domain anchoring distances, which upper bound the semantic distance between the input and the translated features based on the triangle inequality. Fig.~\ref{fig:pa} illustrates the way we perform latent space anchoring. 
Let $x_A$ be an image from domain $A$. Recall Eq. \ref{eq:recon}, $x_A$ is translated to a domain $B$ image $x_{A\to B}=R_B(G_\text{feat}(E_A(x_A)))$. The goal of image translation is to minimize $||f_{{A\to B}} - f_A||$. Because of the triangle inequality,
\begin{equation}
    ||f_{A\to B} - f_A||_2^2  \le ||f_{A\to B} - f_{\star}||_2^2 + ||f_{A} - f_{\star}||_2^2, \label{eq:5}
\end{equation}
where $f^{\star}$ is the shared ideal anchor point in pre-defined generator feature space. Eq. \ref{eq:5} tells that the intractable minimization of the perceptual discrepancy can be transformed into the tractable minimization of $||f - f_{\star}||_2^2$ on two single image domains, individually, as illustrated in Eq. \ref{eq:total_loss}.

\bibliographystyle{IEEEtran}
\bibliography{main}

\begin{thebibliography}{10}
\providecommand{\url}[1]{#1}
\csname url@samestyle\endcsname
\providecommand{\newblock}{\relax}
\providecommand{\bibinfo}[2]{#2}
\providecommand{\BIBentrySTDinterwordspacing}{\spaceskip=0pt\relax}
\providecommand{\BIBentryALTinterwordstretchfactor}{4}
\providecommand{\BIBentryALTinterwordspacing}{\spaceskip=\fontdimen2\font plus
\BIBentryALTinterwordstretchfactor\fontdimen3\font minus
  \fontdimen4\font\relax}
\providecommand{\BIBforeignlanguage}[2]{{%
\expandafter\ifx\csname l@#1\endcsname\relax
\typeout{** WARNING: IEEEtran.bst: No hyphenation pattern has been}%
\typeout{** loaded for the language `#1'. Using the pattern for}%
\typeout{** the default language instead.}%
\else
\language=\csname l@#1\endcsname
\fi
#2}}
\providecommand{\BIBdecl}{\relax}
\BIBdecl

\bibitem{cyclegan}
J.-Y. Zhu, T.~Park, P.~Isola, and A.~A. Efros, ``Unpaired image-to-image
  translation using cycle-consistent adversarial networks,'' in \emph{ICCV},
  2017, pp. 2223--2232.

\bibitem{unit}
M.-Y. Liu, T.~Breuel, and J.~Kautz, ``Unsupervised image-to-image translation
  networks,'' \emph{NeurIPS}, vol.~30, 2017.

\bibitem{cut}
T.~Park, A.~A. Efros, R.~Zhang, and J.-Y. Zhu, ``Contrastive learning for
  unpaired image-to-image translation,'' in \emph{ECCV}, 2020, pp. 319--345.

\bibitem{pix2pix}
P.~Isola, J.-Y. Zhu, T.~Zhou, and A.~A. Efros, ``Image-to-image translation
  with conditional adversarial networks,'' in \emph{CVPR}, 2017, pp.
  1125--1134.

\bibitem{spade}
T.~Park, M.-Y. Liu, T.-C. Wang, and J.-Y. Zhu, ``Semantic image synthesis with
  spatially-adaptive normalization,'' in \emph{ICCV}, 2019, pp. 2337--2346.

\bibitem{drit++}
H.-Y. Lee, H.-Y. Tseng, Q.~Mao, J.-B. Huang, Y.-D. Lu, M.~Singh, and M.-H.
  Yang, ``Drit++: Diverse image-to-image translation via disentangled
  representations,'' \emph{IJCV}, vol. 128, no.~10, pp. 2402--2417, 2020.

\bibitem{stargan}
Y.~Choi, M.~Choi, M.~Kim, J.-W. Ha, S.~Kim, and J.~Choo, ``Stargan: Unified
  generative adversarial networks for multi-domain image-to-image
  translation,'' in \emph{CVPR}, 2018, pp. 8789--8797.

\bibitem{starganv2}
Y.~Choi, Y.~Uh, J.~Yoo, and J.-W. Ha, ``Stargan v2: Diverse image synthesis for
  multiple domains,'' in \emph{CVPR}, 2020, pp. 8188--8197.

\bibitem{stylegan2}
T.~Karras, S.~Laine, M.~Aittala, J.~Hellsten, J.~Lehtinen, and T.~Aila,
  ``Analyzing and improving the image quality of stylegan,'' in \emph{CVPR},
  2020, pp. 8110--8119.

\bibitem{biggan}
A.~Brock, J.~Donahue, and K.~Simonyan, ``Large scale gan training for high
  fidelity natural image synthesis,'' in \emph{ICLR}, 2018.

\bibitem{almahairi2018augmented}
A.~Almahairi, S.~Rajeshwar, A.~Sordoni, P.~Bachman, and A.~Courville,
  ``Augmented cyclegan: Learning many-to-many mappings from unpaired data,'' in
  \emph{ICML}, 2018, pp. 195--204.

\bibitem{kim2019u}
J.~Kim, M.~Kim, H.~Kang, and K.~Lee, ``U-gat-it: Unsupervised generative
  attentional networks with adaptive layer-instance normalization for
  image-to-image translation,'' \emph{arXiv preprint arXiv:1907.10830}, 2019.

\bibitem{zhao2020unpaired}
Y.~Zhao, R.~Wu, and H.~Dong, ``Unpaired image-to-image translation using
  adversarial consistency loss,'' in \emph{ECCV}, 2020, pp. 800--815.

\bibitem{chen2020reusing}
R.~Chen, W.~Huang, B.~Huang, F.~Sun, and B.~Fang, ``Reusing discriminators for
  encoding: Towards unsupervised image-to-image translation,'' in \emph{CVPR},
  2020, pp. 8168--8177.

\bibitem{councilgan}
O.~Nizan and A.~Tal, ``Breaking the cycle-colleagues are all you need,'' in
  \emph{CVPR}, 2020, pp. 7860--7869.

\bibitem{shao2021spatchgan}
X.~Shao and W.~Zhang, ``Spatchgan: A statistical feature based discriminator
  for unsupervised image-to-image translation,'' in \emph{ICCV}, 2021, pp.
  6546--6555.

\bibitem{zhao2021unpaired}
Y.~Zhao and C.~Chen, ``Unpaired image-to-image translation via latent energy
  transport,'' in \emph{CVPR}, 2021, pp. 16\,418--16\,427.

\bibitem{pizzati2021comogan}
F.~Pizzati, P.~Cerri, and R.~de~Charette, ``Comogan: continuous model-guided
  image-to-image translation,'' in \emph{CVPR}, 2021, pp. 14\,288--14\,298.

\bibitem{discogan}
T.~Kim, M.~Cha, H.~Kim, J.~K. Lee, and J.~Kim, ``Learning to discover
  cross-domain relations with generative adversarial networks,'' in
  \emph{ICML}, 2017, pp. 1857--1865.

\bibitem{dualgan}
Z.~Yi, H.~Zhang, P.~Tan, and M.~Gong, ``Dualgan: Unsupervised dual learning for
  image-to-image translation,'' in \emph{ICCV}, 2017, pp. 2849--2857.

\bibitem{liu2020gmm}
Y.~Liu, M.~De~Nadai, J.~Yao, N.~Sebe, B.~Lepri, and X.~Alameda-Pineda,
  ``Gmm-unit: Unsupervised multi-domain and multi-modal image-to-image
  translation via attribute gaussian mixture modeling,'' \emph{arXiv preprint
  arXiv:2003.06788}, 2020.

\bibitem{xu2021domain}
W.~Xu and G.~Wang, ``A domain gap aware generative adversarial network for
  multi-domain image translation,'' \emph{IEEE TIP}, vol.~31, pp. 72--84, 2021.

\bibitem{vinod2021multi}
V.~Vinod, K.~R. Prabhakar, R.~V. Babu, and A.~Chakraborty, ``Multi-domain
  conditional image translation: Translating driving datasets from
  clear-weather to adverse conditions,'' in \emph{ICCV}, 2021, pp. 1571--1582.

\bibitem{munit}
X.~Huang, M.-Y. Liu, S.~Belongie, and J.~Kautz, ``Multimodal unsupervised
  image-to-image translation,'' in \emph{ECCV}, 2018, pp. 172--189.

\bibitem{drit}
H.-Y. Lee, H.-Y. Tseng, J.-B. Huang, M.~Singh, and M.-H. Yang, ``Diverse
  image-to-image translation via disentangled representations,'' in
  \emph{ECCV}, 2018, pp. 35--51.

\bibitem{zhao2018modular}
B.~Zhao, B.~Chang, Z.~Jie, and L.~Sigal, ``Modular generative adversarial
  networks,'' in \emph{ECCV}, 2018, pp. 150--165.

\bibitem{pumarola2018ganimation}
A.~Pumarola, A.~Agudo, A.~M. Martinez, A.~Sanfeliu, and F.~Moreno-Noguer,
  ``Ganimation: Anatomically-aware facial animation from a single image,'' in
  \emph{ECCV}, 2018, pp. 818--833.

\bibitem{funit}
M.-Y. Liu, X.~Huang, A.~Mallya, T.~Karras, T.~Aila, J.~Lehtinen, and J.~Kautz,
  ``Few-shot unsupervised image-to-image translation,'' in \emph{ICCV}, 2019,
  pp. 10\,551--10\,560.

\bibitem{romero2019smit}
A.~Romero, P.~Arbel{\'a}ez, L.~Van~Gool, and R.~Timofte, ``Smit: Stochastic
  multi-label image-to-image translation,'' in \emph{ICCV Workshops}, 2019, pp.
  0--0.

\bibitem{wang2019sdit}
Y.~Wang, A.~Gonzalez-Garcia, J.~van~de Weijer, and L.~Herranz, ``Sdit: Scalable
  and diverse cross-domain image translation,'' in \emph{Proceedings of the
  27th ACM International Conference on Multimedia}, 2019, pp. 1267--1276.

\bibitem{chen2020domain}
Y.-C. Chen, X.~Xu, and J.~Jia, ``Domain adaptive image-to-image translation,''
  in \emph{CVPR}, 2020, pp. 5274--5283.

\bibitem{tunit}
K.~Baek, Y.~Choi, Y.~Uh, J.~Yoo, and H.~Shim, ``Rethinking the truly
  unsupervised image-to-image translation,'' in \emph{ICCV}, 2021, pp.
  14\,154--14\,163.

\bibitem{liu2021smoothing}
Y.~Liu, E.~Sangineto, Y.~Chen, L.~Bao, H.~Zhang, N.~Sebe, B.~Lepri, W.~Wang,
  and M.~De~Nadai, ``Smoothing the disentangled latent style space for
  unsupervised image-to-image translation,'' in \emph{CVPR}, 2021, pp.
  10\,785--10\,794.

\bibitem{pang2021image}
Y.~Pang, J.~Lin, T.~Qin, and Z.~Chen, ``Image-to-image translation: Methods and
  applications,'' \emph{IEEE Transactions on Multimedia}, 2021.

\bibitem{combogan}
A.~Anoosheh, E.~Agustsson, R.~Timofte, and L.~Van~Gool, ``Combogan:
  Unrestrained scalability for image domain translation,'' in \emph{CVPR
  Workshops}, 2018, pp. 783--790.

\bibitem{gandissection}
D.~Bau, J.-Y. Zhu, H.~Strobelt, B.~Zhou, J.~B. Tenenbaum, W.~T. Freeman, and
  A.~Torralba, ``Gan dissection: Visualizing and understanding generative
  adversarial networks,'' in \emph{ICLR}, 2019.

\bibitem{xia2021gan}
W.~Xia, Y.~Zhang, Y.~Yang, J.-H. Xue, B.~Zhou, and M.-H. Yang, ``Gan inversion:
  A survey,'' \emph{arXiv preprint arXiv:2101.05278}, 2021.

\bibitem{image2stylegan}
R.~Abdal, Y.~Qin, and P.~Wonka, ``Image2stylegan: How to embed images into the
  stylegan latent space?'' in \emph{ICCV}, 2019, pp. 4432--4441.

\bibitem{psp}
E.~Richardson, Y.~Alaluf, O.~Patashnik, Y.~Nitzan, Y.~Azar, S.~Shapiro, and
  D.~Cohen-Or, ``Encoding in style: a stylegan encoder for image-to-image
  translation,'' in \emph{CVPR}, 2021, pp. 2287--2296.

\bibitem{shen2020interpreting}
Y.~Shen, J.~Gu, X.~Tang, and B.~Zhou, ``Interpreting the latent space of gans
  for semantic face editing,'' in \emph{CVPR}, 2020, pp. 9243--9252.

\bibitem{cherepkov2021navigating}
A.~Cherepkov, A.~Voynov, and A.~Babenko, ``Navigating the gan parameter space
  for semantic image editing,'' in \emph{CVPR}, 2021, pp. 3671--3680.

\bibitem{tov2021designing}
O.~Tov, Y.~Alaluf, Y.~Nitzan, O.~Patashnik, and D.~Cohen-Or, ``Designing an
  encoder for stylegan image manipulation,'' \emph{ACM TOG}, vol.~40, no.~4,
  pp. 1--14, 2021.

\bibitem{dgp}
X.~Pan, X.~Zhan, B.~Dai, D.~Lin, C.~C. Loy, and P.~Luo, ``Exploiting deep
  generative prior for versatile image restoration and manipulation,''
  \emph{IEEE TPAMI}, 2021.

\bibitem{image2stylegan++}
R.~Abdal, Y.~Qin, and P.~Wonka, ``Image2stylegan++: How to edit the embedded
  images?'' in \emph{CVPR}, 2020, pp. 8296--8305.

\bibitem{shen2021closed}
Y.~Shen and B.~Zhou, ``Closed-form factorization of latent semantics in gans,''
  in \emph{CVPR}, 2021, pp. 1532--1540.

\bibitem{voynov2020unsupervised}
A.~Voynov and A.~Babenko, ``Unsupervised discovery of interpretable directions
  in the gan latent space,'' in \emph{ICML}, 2020, pp. 9786--9796.

\bibitem{datasetgan}
Y.~Zhang, H.~Ling, J.~Gao, K.~Yin, J.-F. Lafleche, A.~Barriuso, A.~Torralba,
  and S.~Fidler, ``Datasetgan: Efficient labeled data factory with minimal
  human effort,'' in \emph{CVPR}, 2021, pp. 10\,145--10\,155.

\bibitem{segmentationinstyle}
D.~Pakhomov, S.~Hira, N.~Wagle, K.~E. Green, and N.~Navab, ``Segmentation in
  style: Unsupervised semantic image segmentation with stylegan and clip,''
  \emph{arXiv preprint arXiv:2107.12518}, 2021.

\bibitem{benaim2018one}
S.~Benaim and L.~Wolf, ``One-shot unsupervised cross domain translation,''
  \emph{advances in neural information processing systems}, vol.~31, 2018.

\bibitem{back2021fine}
J.~Back, ``Fine-tuning stylegan2 for cartoon face generation,'' \emph{arXiv
  preprint arXiv:2106.12445}, 2021.

\bibitem{freezeg}
S.~Mo, M.~Cho, and J.~Shin, ``Freeze the discriminator: a simple baseline for
  fine-tuning gans,'' \emph{arXiv preprint arXiv:2002.10964}, 2020.

\bibitem{stylegan3}
K.~et~al., ``Alias-free generative adversarial networks,'' in \emph{NeurIPS},
  2021.

\bibitem{stylegan}
T.~Karras, S.~Laine, and T.~Aila, ``A style-based generator architecture for
  generative adversarial networks,'' in \emph{CVPR}, 2019, pp. 4401--4410.

\bibitem{celebamaskhq}
C.-H. Lee, Z.~Liu, L.~Wu, and P.~Luo, ``Maskgan: Towards diverse and
  interactive facial image manipulation,'' in \emph{CVPR}, 2020, pp.
  5549--5558.

\bibitem{cufsf1}
X.~Wang and X.~Tang, ``Face photo-sketch synthesis and recognition,''
  \emph{IEEE TPAMI}, vol.~31, no.~11, pp. 1955--1967, 2008.

\bibitem{cufsf2}
W.~Zhang, X.~Wang, and X.~Tang, ``Coupled information-theoretic encoding for
  face photo-sketch recognition,'' in \emph{CVPR}, 2011, pp. 513--520.

\bibitem{landmark}
V.~Kazemi and J.~Sullivan, ``One millisecond face alignment with an ensemble of
  regression trees,'' in \emph{CVPR}, 2014, pp. 1867--1874.

\bibitem{king2009dlib}
D.~E. King, ``Dlib-ml: A machine learning toolkit,'' \emph{JMLR}, vol.~10, pp.
  1755--1758, 2009.

\bibitem{imagenet}
J.~Deng, W.~Dong, R.~Socher, L.-J. Li, K.~Li, and L.~Fei-Fei, ``Imagenet: A
  large-scale hierarchical image database,'' in \emph{CVPR}, 2009, pp.
  248--255.

\bibitem{imagenetseg1}
D.~Kuettel, M.~Guillaumin, and V.~Ferrari, ``Segmentation propagation in
  imagenet,'' in \emph{ECCV}, 2012, pp. 459--473.

\bibitem{imagenetseg2}
M.~Guillaumin, D.~K{\"u}ttel, and V.~Ferrari, ``Imagenet auto-annotation with
  segmentation propagation,'' \emph{IJCV}, vol. 110, no.~3, pp. 328--348, 2014.

\bibitem{resnet}
K.~He, X.~Zhang, S.~Ren, and J.~Sun, ``Deep residual learning for image
  recognition,'' in \emph{CVPR}, 2016, pp. 770--778.

\bibitem{adam}
D.~P. Kingma and J.~Ba, ``Adam: A method for stochastic optimization,'' in
  \emph{ICLR}, 2015.

\bibitem{fid}
M.~Heusel, H.~Ramsauer, T.~Unterthiner, B.~Nessler, and S.~Hochreiter, ``Gans
  trained by a two time-scale update rule converge to a local nash
  equilibrium,'' \emph{NeurIPS}, vol.~30, 2017.

\bibitem{lpips}
R.~Zhang, P.~Isola, A.~A. Efros, E.~Shechtman, and O.~Wang, ``The unreasonable
  effectiveness of deep features as a perceptual metric,'' in \emph{CVPR},
  2018, pp. 586--595.

\bibitem{freezesg}
J.~Back, ``Fine-tuning stylegan2 for cartoon face generation,'' \emph{arXiv
  preprint arXiv:2106.12445}, 2021.

\bibitem{inversion}
J.~Zhu, Y.~Shen, D.~Zhao, and B.~Zhou, ``In-domain gan inversion for real image
  editing,'' in \emph{ECCV}, 2020.

\bibitem{jaderberg2015spatial}
M.~Jaderberg, K.~Simonyan, A.~Zisserman \emph{et~al.}, ``Spatial transformer
  networks,'' \emph{NeurIPS}, vol.~28, 2015.

\bibitem{dai2017deformable}
J.~Dai, H.~Qi, Y.~Xiong, Y.~Li, G.~Zhang, H.~Hu, and Y.~Wei, ``Deformable
  convolutional networks,'' in \emph{ICCV}, 2017, pp. 764--773.

\bibitem{zheng2022semantic}
H.~Zheng, Z.~Lin, J.~Lu, S.~Cohen, J.~Zhang, N.~Xu, and J.~Luo, ``Semantic
  layout manipulation with high-resolution sparse attention,'' \emph{IEEE
  Transactions on Pattern Analysis and Machine Intelligence}, 2022.

\bibitem{zheng2022cm}
H.~Zheng, Z.~Lin, J.~Lu, S.~Cohen, E.~Shechtman, C.~Barnes, J.~Zhang, N.~Xu,
  S.~Amirghodsi, and J.~Luo, ``Cm-gan: Image inpainting with cascaded
  modulation gan and object-aware training,'' 2022.

\bibitem{tan2020crossnet++}
Y.~Tan, H.~Zheng, Y.~Zhu, X.~Yuan, X.~Lin, D.~Brady, and L.~Fang, ``Crossnet++:
  Cross-scale large-parallax warping for reference-based super-resolution,''
  \emph{IEEE TPAMI}, vol.~43, no.~12, pp. 4291--4305, 2020.

\bibitem{zhang2020cross}
P.~Zhang, B.~Zhang, D.~Chen, L.~Yuan, and F.~Wen, ``Cross-domain correspondence
  learning for exemplar-based image translation,'' in \emph{CVPR}, 2020, pp.
  5143--5153.

\end{thebibliography}

% \ifCLASSOPTIONcaptionsoff
%   \newpage
% \fi

\begin{IEEEbiography}[{\includegraphics[width=1in,height=1.25in,clip,keepaspectratio]{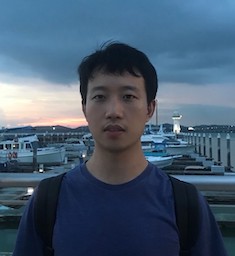}}]{Siyu Huang}
received the B.E. degree and Ph.D. degree in information and communication engineering from Zhejiang University, Hangzhou, China, in 2014 and 2019. He is currently a Postdoctoral Fellow in the John A. Paulson School of Engineering and Applied Sciences at Harvard University. Before that, he was a Visiting Scholar at Language Technologies Institute in the School of Computer Science, Carnegie Mellon University in 2018, a Research Scientist at Big Data Laboratory, Baidu Research from 2019 to 2021, and a Research Fellow in the School of Electrical and Electronic Engineering at Nanyang Technological University in 2021. He has published more than 20 papers on top-tier computer science journals and conferences. His research interests are primarily in computer vision, multimedia analysis, and generative models.
\end{IEEEbiography}

\begin{IEEEbiography}[{\includegraphics[width=1in,height=1.25in,clip,keepaspectratio]{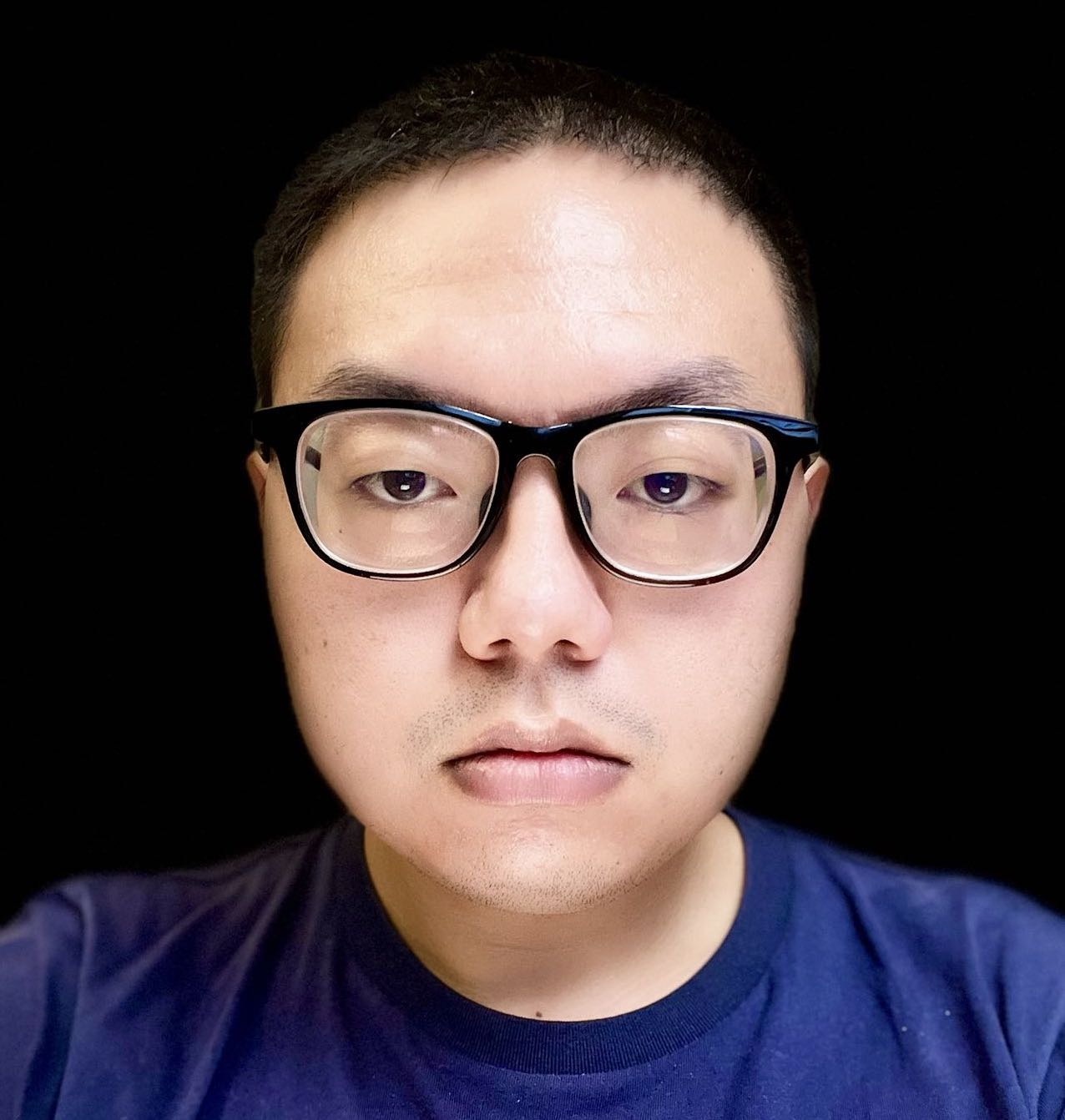}}]{Jie An}
received his Bachelor of Science (B.Sc) in Information and Computing Sciences and Master of Science (M.Sc) in Applied Mathematics, both from Peking University. He is currently a PhD student in Computer Science at the University of Rochester. His research interests include computer Vision, generative models, and AI+Art. 
\end{IEEEbiography}

\begin{IEEEbiography}[{\includegraphics[width=1in,height=1.25in,clip,keepaspectratio]{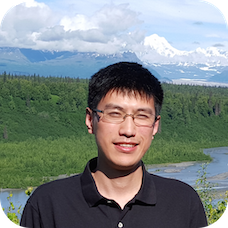}}]{Donglai Wei}
received his Bachelor of Science (B.Sc) in Mathematics from Brown University and PhD degree in Computer Science from MIT. He is an assistant professor of computer science at Boston College. Before joining Boston College, he worked as a postdoctoral researcher at Harvard University. His research interests include brain image analysis and video understanding.
\end{IEEEbiography}

\begin{IEEEbiography}[{\includegraphics[width=1in,height=1.25in,clip,keepaspectratio]{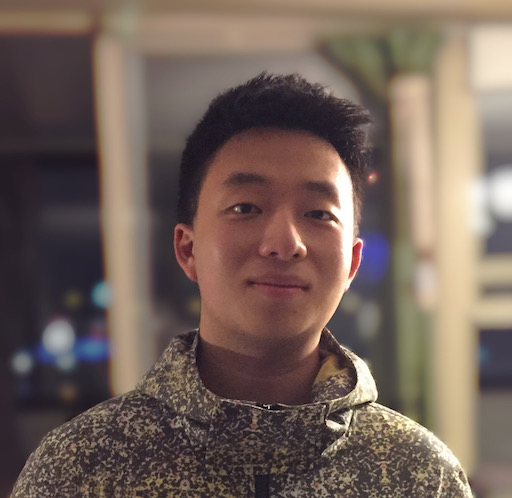}}]{Zudi Lin}
received his PhD degree in Computer Science in 2022 from the John A. Paulson School of Engineering and Applied Sciences at Harvard Univerisity. Before that, he obtained M.S. in Computer Science from Harvard University in 2020 and B.S. in Biological Science from Tsinghua University in 2017. His research interests include deep learning, computer vision, and neuroscience.
\end{IEEEbiography}

\begin{IEEEbiography}[{\includegraphics[width=1in,height=1.25in,clip,keepaspectratio]{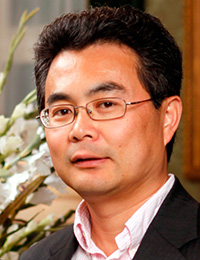}}]{Jiebo Luo} (Fellow, IEEE)
is the Albert Arendt Hopeman Professor of Engineering and Professor of Computer Science at the University of Rochester which he joined in 2011 after a
prolific career of fifteen years at Kodak Research Laboratories. He has authored nearly 600 technical papers and holds over 90 U.S. patents. His research interests include computer vision, NLP, machine learning, data mining, computational social science, and digital health. He has been
involved in numerous technical conferences, including serving as program co-chair of ACM
Multimedia 2010, IEEE CVPR 2012, ACM ICMR 2016, and IEEE ICIP 2017, as well as general co-chair of ACM Multimedia 2018 and IEEE ICME 2024. He has served on the editorial boards of the IEEE Transactions on Pattern
Analysis and Machine Intelligence (TPAMI), IEEE Transactions on Multimedia (TMM), IEEE Transactions on Circuits and Systems for Video Technology (TCSVT), IEEE Transactions on Big Data (TBD), ACM Transactions on Intelligent Systems and Technology (TIST), Pattern
Recognition, Knowledge and Information Systems (KAIS), Machine Vision and Applications, and Intelligent Medicine. He was the Editor-in-Chief of the IEEE Transactions on Multimedia (2020-2022). Professor Luo is also a Fellow of NAI, ACM, AAAI, SPIE, and IAPR.
\end{IEEEbiography}

\begin{IEEEbiography}[{\includegraphics[width=1in,height=1.25in,clip,keepaspectratio]{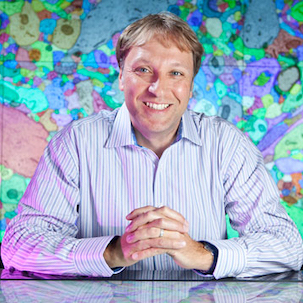}}]{Hanspeter Pfister} (Fellow, IEEE) received the MS degree in electrical engineering from ETH Zurich, Switzerland, and the PhD degree in computer science from the State University of New York at Stony Brook. He is currently a professor of computer science at the Harvard John A. Paulson School of Engineering and Applied Sciences and an affiliate faculty member of the Center for Brain Science. His research in visual computing lies at the intersection of visualization, computer graphics, and computer vision and spans a wide range of topics, including biomedical image analysis and visualization, image and video analysis, interpretable machine learning, and visual analytics in data science. From 2013 to 2017, he was the director of the Institute for Applied Computational Science. Before joining Harvard, he worked for over a decade at Mitsubishi Electric Research Laboratories, where he was an associate director and senior research scientist. He was the chief architect of VolumePro, Mitsubishi Electric’s award-winning real-time volume rendering graphics card, for which he received the Mitsubishi Electric President’s Award, in 2000. He was elected as an ACM fellow, in 2019, and an IEEE fellow, in 2022.
\end{IEEEbiography}

\end{document}